\pdfoutput=1

\documentclass[11pt]{article}

\usepackage{EMNLP2022}

\usepackage{times}
\usepackage{latexsym}

\usepackage[T1]{fontenc}

\usepackage[utf8]{inputenc}

\usepackage{microtype}

\usepackage{inconsolata}
\usepackage{graphicx}
\usepackage{subcaption}
\usepackage{amsmath}
\usepackage{amsfonts,amssymb}
\usepackage{booktabs}
\usepackage{enumerate}
\usepackage{enumitem}
\usepackage{multirow}
\usepackage{lipsum}
\usepackage{pifont}
\usepackage{url}
\usepackage{float}
\newcommand{\xmark}{\ding{55}}
\newcommand{\greencheck}{{\color{green}\checkmark}}
\newcommand{\redx}{{\color{red}\xmark}}
\newcommand{\head}[1]{{\textbf{#1}}}

\title{AssistSR: Task-oriented Video Segment Retrieval for Personal AI Assistant}

\author{
Stan Weixian Lei, Difei Gao, Yuxuan Wang, Dongxing Mao, \\
\textbf{Zihan Liang, Lingmin Ran, Mike Zheng Shou}$^{\dag}$\\
 \\
Show Lab, National University of Singapore
}

\begin{document}
\maketitle
\def\thefootnote{$^{\dag}$}\footnotetext{Corresponding author.}\def\thefootnote{\arabic{footnote}}
\begin{abstract}
It is still a pipe dream that personal AI assistants on the phone and AR glasses can assist our daily life in addressing our questions like ``how to adjust the date for this watch?'' and ``how to set its heating duration? (while pointing at an oven)''. The queries used in conventional tasks (i.e. Video Question Answering, Video Retrieval, Moment Localization) are often factoid and based on pure text. In contrast, we present a new task called Task-oriented Question-driven Video Segment Retrieval (TQVSR). Each of our questions is an image-box-text query that focuses on affordance of items in our daily life and expects relevant answer segments to be retrieved from a corpus of instructional video-transcript segments. To support the study of this TQVSR task, we construct a new dataset called AssistSR. We design novel guidelines to create high-quality samples. This dataset contains 3.2k multimodal questions on 1.6k video segments from instructional videos on diverse daily-used items. To address TQVSR, we develop a simple yet effective model called Dual Multimodal Encoders (DME) that significantly outperforms several baseline methods while still having large room for improvement in the future. Moreover, we present detailed ablation analyses.
Code and data are available at \url{https://github.com/StanLei52/TQVSR}.
\end{abstract}

\section{Introduction}

\begin{figure}[t]
  \centering
  \begin{subfigure}{0.99\linewidth}
\includegraphics[width=0.99\linewidth]{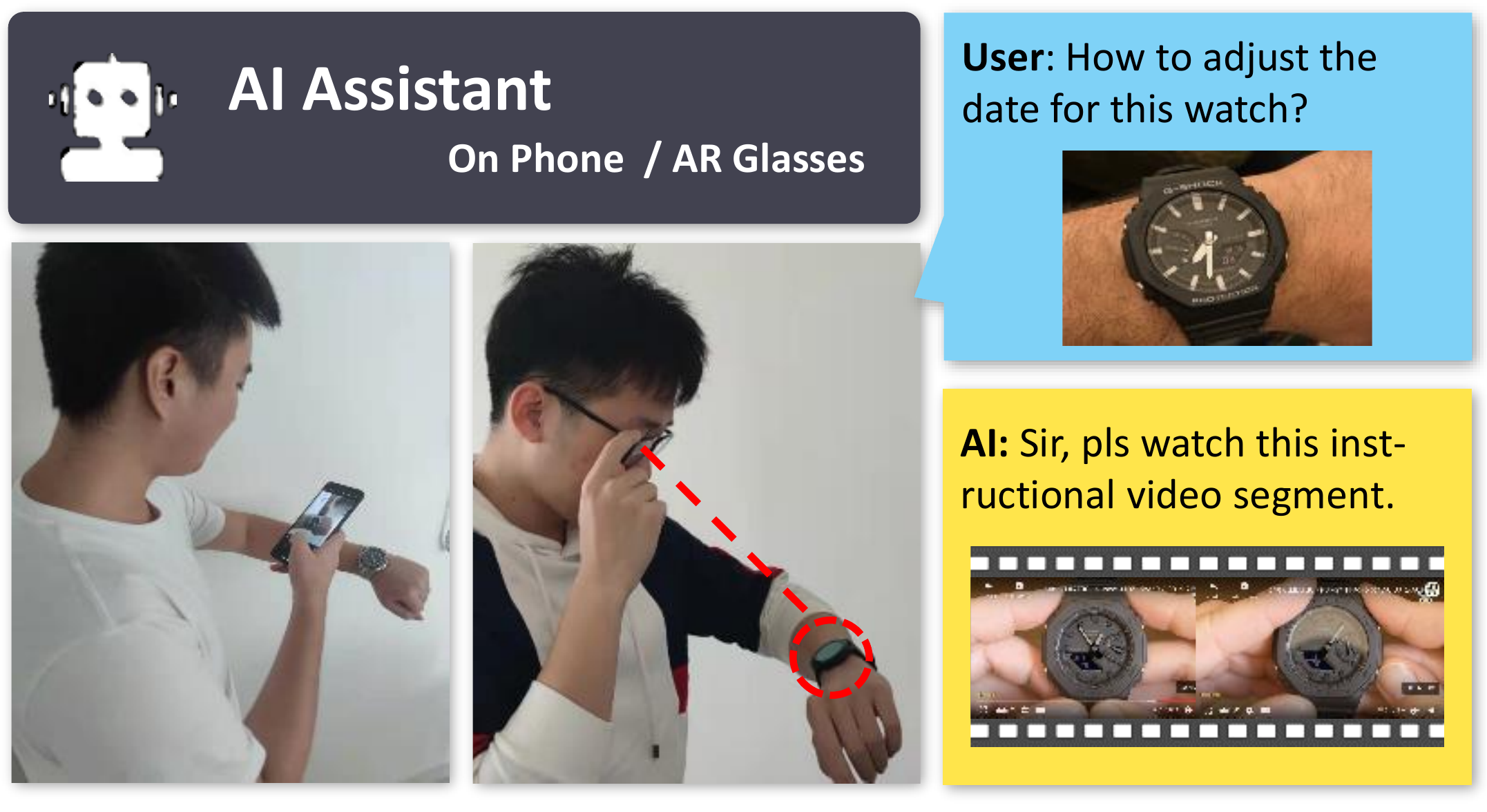}

    \caption{ }
    \label{fig:app}
  \end{subfigure}
  \hfill
  
  \begin{subfigure}{0.99\linewidth}
\includegraphics[width=0.99\linewidth]{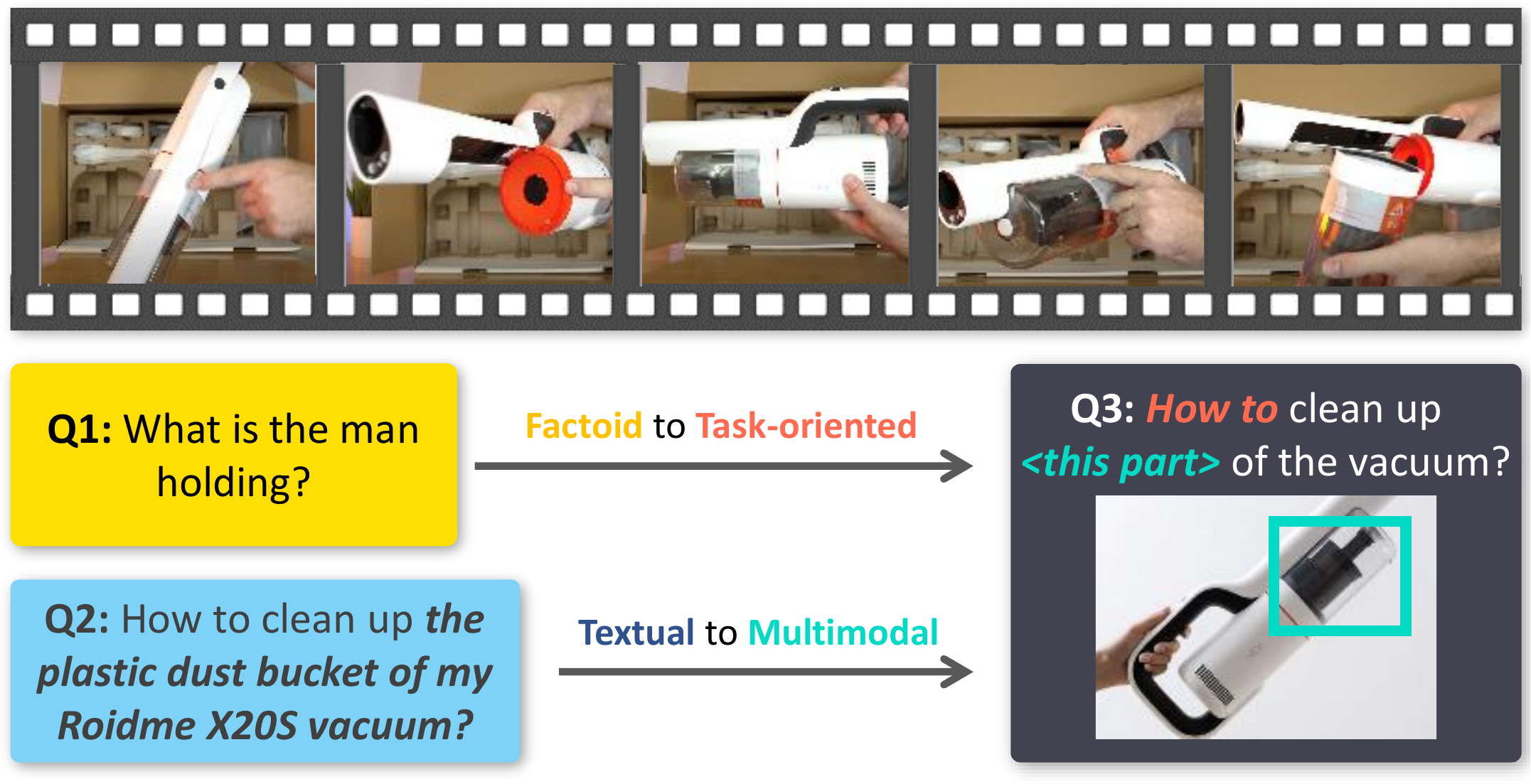}

    \caption{ }
    \label{fig:questions_cmp}
  \end{subfigure}
  \caption{(a) AI assistant on phone/glasses addresses user's question in daily life. (b) Questions in previous tasks vs. our TQVSR task.}
  \label{fig:teaser}
\end{figure}

As shown in Fig.~\ref{fig:app}, a user is \textit{looking at his watch and wondering ``how to adjust the date for this watch"}.
It would be great if our phone or glasses can be powered by an intelligent agent, i.e. AI assistant, \textit{which perceives exactly what the user sees and can find out the answer in the instructional video of this watch}, either provided by the watch's seller or posted on YouTube.
However, such a video is usually quite long and thus time-consuming for users to watch, let alone there are a lot of online tutorial videos for various items. 
To save a user's time, we aim to retrieve only the relevant short segments in a long video for that specific watch. 

\definecolor{color_inp}{RGB}{174,213,250}
\definecolor{color_source}{RGB}{254,249,184}
\definecolor{color_outp}{RGB}{212, 234, 188}

\begin{figure*}[th]
  \centering
  \includegraphics[width=0.95\textwidth]{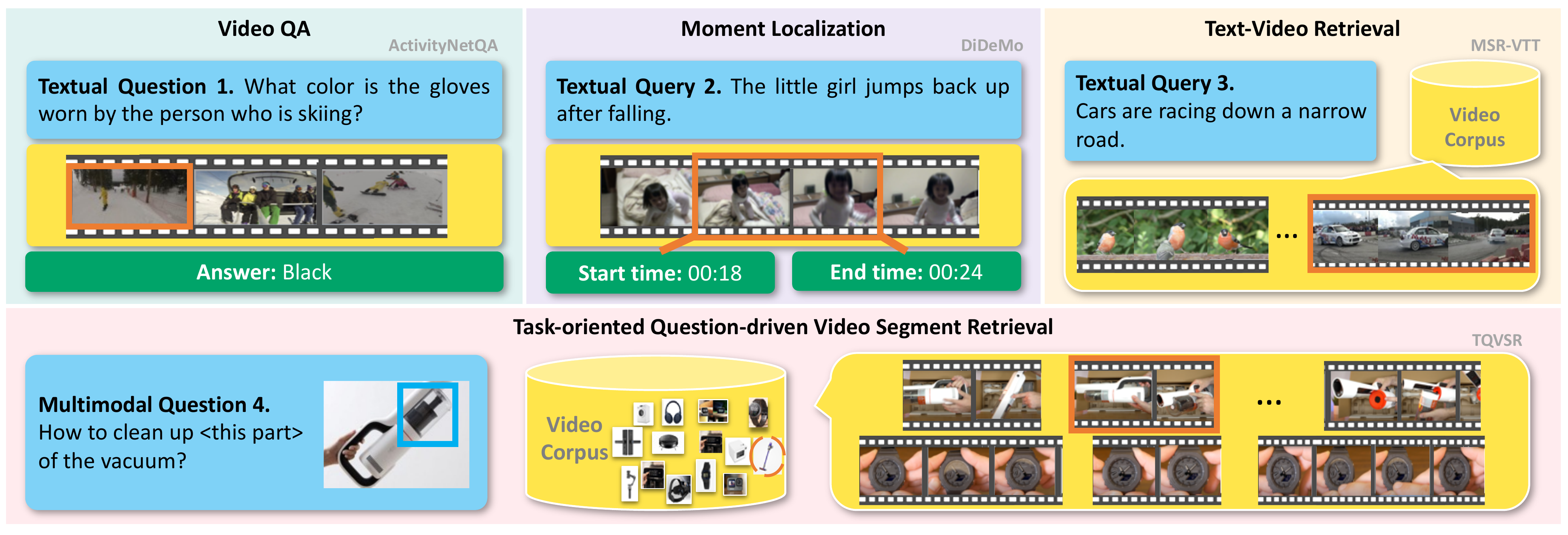}
  \caption{
  Illustration for TQVSR \&  TQVSR \textbf{vs.} other video-related tasks. 
  TQVSR focuses on task-oriented and multimodal questions: an image or referring regions as visual part and a natural language query as textual part. TQVSR is set-based: given a question, the system is to retrieve from the video corpus the relevant segments.
  }
  \label{fig:task}
\end{figure*}

This problem is related with several existing tasks including Video Question Answering (VideoQA), Video Retrieval and Moment Localization. Yet, to unlock the aforementioned application in Fig.~\ref{fig:app}, we need a new task and a new benchmark, which have two major differences compared to the existing tasks when designing \textbf{questions}:
\begin{itemize}[leftmargin=12pt]
\setlength{\itemsep}{2pt}
\setlength{\parsep}{0pt}
\setlength{\parskip}{0pt}
    \item \textbf{Factoid vs. Task-oriented.} 
    Existing tasks and datasets mainly focus on factoid questions or queries. For example in Fig.~\ref{fig:task}, Question 1 in ActivityNetQA ~\cite{yu2019activitynetQA}, Query 2 in DiDeMo dataset~\cite{anne2017localizing}, and Query 3 in MSR-VTT~\cite{xu2016msr-vtt} are all about simple facts, e.g. basic attributes, relationships between common objects. In contrast, \textit{we expect the AI assistant to go beyond simple facts and tackle Task-oriented questions}. 
    As shown in Fig.~\ref{fig:questions_cmp}, instead of asking ``what is the man holding'', we focus on questions like ``how to clean up \texttt{<this part>} of the vaccum''. 
    Such questions regarding the affordance~\cite{gibson1977theory} of objects or devices are often asked by people in daily life.

    \item \textbf{Pure text vs. Multimodal.} Questions or queries in existing datasets are mostly pure text. However, psychology literature shows that pointing to the interesting objects is one of the initial manners that a baby conveys intention \cite{mani2020point, oates2004cognitive,malle2001intentions}. It is unnatural for human to ask the AI assistant with phrases like Q2 in Fig~\ref{fig:questions_cmp} or ``on the top-right of the vacuum''. 
    It is more straightforward to directly point to that part and ask: ``How to clean up \colorbox{color_inp}{\texttt{<this part\includegraphics[scale=.07, bb=-50 25 150 14]{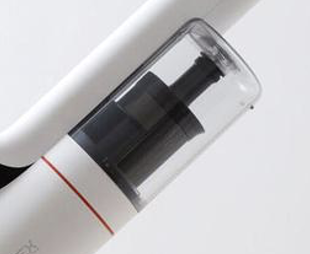}>}} of the vacuum?''. 
    This is a multimodal question consisting of \textit{a textual question, the current image seen by the AI assistant, and the visual region pointed out by the user's finger.}
\end{itemize}


\textbf{Task}. Given a multimodal, task-oriented question, we propose a new task called Task-oriented Question-driven Video Segment Retrieval (\textbf{TQVSR}), which expects the model to retrieve the relevant video segments that can address the user's question. 
As shown in Fig.~\ref{fig:task}, (1) unlike video-text retrieval which focuses on returning a whole video, TQVSR takes a step further to locate the relevant segment within a long instructional video. (2) Unlike moment localization whose query directly describes the segment's content, we find it is often difficult for different annotators to agree on where the exact start and end times are for task-oriented question. To provide a fair comparisons, TQVSR first asks annotators to chunk a long video into short segments and then models can only focus on whether retrieve a segment or not. Details in Sec.~\ref{sec:video_seg}.

\textbf{Dataset}. To support studying TQVSR, we create a new dataset called \textbf{AssistSR}. 
Throughout multiple iterations, we form a novel annotation pipeline and guideline, which can effectively address annotation challenges including (1) source high-quality videos that support asking task-oriented questions; (2) annotators tend to ask factoid and simple questions; (3) how to alleviate ambiguity for answer segment annotation. 
In total, we have collected 3,214 high-quality questions on 1,607 segments from 210 videos.

\textbf{Model}. To develop model for TQVSR, we propose Dual Multimodal Encoders (\textbf{DME}), a straightforward yet effective approach that has two transformer-based multi-modal encoders for the user question and a video segment respectively. We further show DME outperforms previous methods and each modality in question and video matters.

\section{Related Work}
\noindent\head{Video Question Answering.} This task requires the intelligent system to automatically answer a natural language question according to the content of a given video. In recent years, multiple VideoQA datasets and tasks have been proposed to facilitate research towards this goal, several video-based QA datasets have also been proposed, e.g. TGIF-QA~\cite{jang2017tgif}, MovieFIB ~\cite{maharaj2017dataset}, VideoQA~\cite{zhu2017uncovering}, LSMDC~\cite{rohrbach2015dataset}, TRECVID~\cite{over2014trecvid}, and MarioQA~\cite{mun2017marioqa}. Additionally, 
Video Question Answering~\cite{lei2018tvqa,tapaswi2016movieqa,kim2017deepstory}, with naturally occurring subtitles are particularly interesting, as it combines both visual and textual information for question answering. Different from VideoQA, where a system is required to generate an answer in a video-instance based setting, the task of TQVSR is to ground the answer segments from the video corpus for a given question.

\noindent\head{Video Retrieval.} The objective of Video Retrieval is that given a text query and a pool of candidate videos, select the video which corresponds to the text query. A good amount of work has been done in the area of natural language query based video search for complete videos, such as MSR-VTT~\cite{xu2016msr-vtt}, DiDeMo~\cite{anne2017localizing}, ActivityNet Captions~\cite{krishna2017dense}, CharadesSTA~\cite{gao2017tall}, and TACoS~\cite{regneri2013grounding}. However, returning the whole video is not always desirable, since sometimes they can be quite long (e.g. from a few minutes to hours). What's more, most of the queries for Video Retrieval are based on declarative sentences, but not question-based, thus leading to the gap to real world application. 
However, the setting of video retrieval is far from practical application due to the factoid query. In TQVSR, we focus more on the task-oriented user question expressed in a multimodal manner and adopt the video segment level retrieval, whereby the returned video segments within proper time-span are able to address the user’s question.




\noindent\head{Moment Localization.} The task of Moment Localization is to localize moments from a video given a natural language query. Various datasets~\cite{anne2017localizing,gao2017tall,lei2020tvr,krishna2017dense,regneri2013grounding} have been proposed or repurposed for the task. While these datasets for moment localization collect only a single moment for each query-video pair, we annotate one or more video segments for each query in our dataset, which is more flexible. Meanwhile, most existing datasets~\cite{sun2014ranking,micheal2016video2gif,song2016click,garcia2018phdgif} for moment localization are query-agnostic, which do not provide customized moments for a specific user query. Recently \cite{lei2021qvhighlights} introduced a dataset for query-based video moment retrieval and highlight detection, which also lies in the category of moment localization. Unlike moment localization, TQVSR is question-driven and aims to retrieve all relevant segments.
Sec.\ref{sec:video_seg} analyzes the challenges in adapting moment localization for TQVSR and our rationales behind formulating it as a segment retrieval problem.

\section{Task Formulation for TQVSR}
\label{sec:TQVSR_formulation}

\label{sec:video_seg}
\noindent\head{Challenges.} Given a user question, naturally we would think of annotating the start time and end time for the answer span, as moment localization does. However, this leads to some problems. (1) \textbf{From the perspective of application}, different users have different preference for answer clips: some prefer to longer clips to learn more contexts, while others just tend to watch the key parts. This leads to the difficulty in the definition of correct answer span. (2) \textbf{From the perspective of metric evaluation}, in the task of moment localization, models are forced to generate a prediction with high tIoU to the ground truth annotation. However, tIoU might not measure the correctness of the predicted answer span well. For example, a predicted answer span with higher tIoU might miss some vital information for addressing the user's question, while another candidate answer span with lower tIoU to ground truth might work better qualitatively. What’s more, a prediction span generated by localization might not be semantically intact, which is not user friendly.

\noindent\textbf{Our proposed solution of evaluating on segments}. To avoid such drawbacks and remove ambiguities in answer annotation, we propose to chunk a video into segments with well-designed rules and then annotate the answer segments among the candidate video segments. With predefined segments, the model no longer needs to predict the timestamp, but only needs to predict whether a segment contains the content for answering the user's question or not, which significantly alleviates ambiguity during evaluation.

Notably, throughout multiple iterations of improving annotation guidelines, we arrived at the following design principles of how to segment a video: 
\textbf{(1)} \textit{Keep consistent level of semantic granularity within one video segment.} 
\textbf{(2)} \textit{Soft threshold for video segment duration: 30 seconds as minimum and 2 minutes as maximum.} 
\textbf{(3)} \textit{Adjust segmentation according to the list of qualified questions.} 
More details and examples could be found in appendix.

\noindent\head{Task definition of TQVSR.} 
The task of TQVSR addresses an task-oriented user question by retrieving from the video corpus the relevant video segments. Taking a question $Q$ with multiple modalities as input, from the corpus of video segments $\mathcal{V}$, the task is to retrieve all the relevant segments $V^Q_{ans}$ which can provide answer contents for the given question: $V_{ans}^Q = \left\{ v_i^q \mid q=Q, i \in \left| \mathcal{V}\right|, v_i^q \in \mathcal{V} \right\}$.

\section{A New Benchmark: AssistSR}
\begin{figure}[t]
\centering
\includegraphics[width=.99\linewidth]{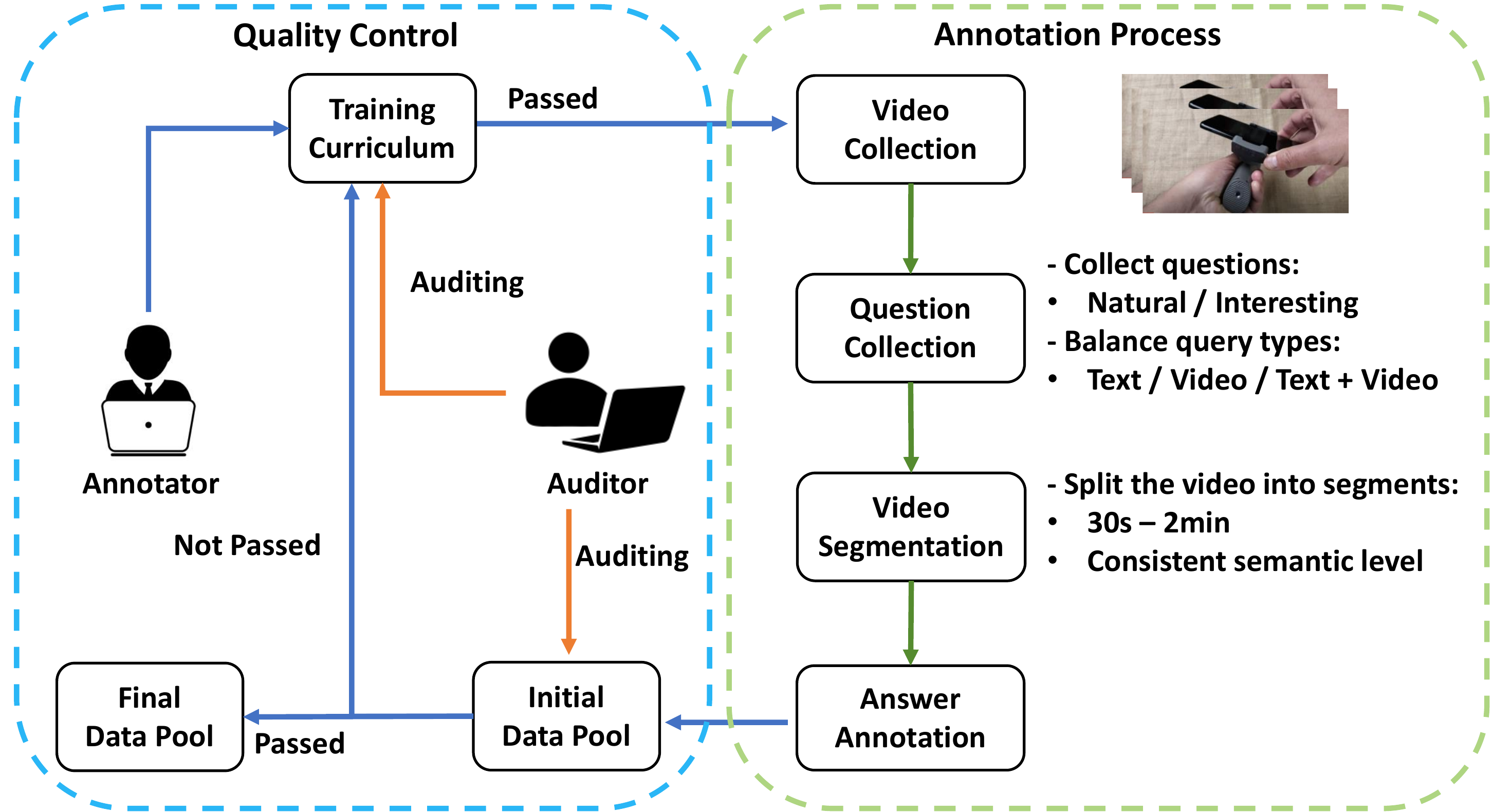}
\caption{Data collection procedure for AssistSR.}
\label{fig:anno_process}
\end{figure}

\head{Overview.} Our pipeline of data collection and annotation procedure is summarized in Fig.~\ref{fig:anno_process}. In short, we conduct video collection, question collection, video segmentation and answer segment annotation for data collection. Quality control is carefully designed to ensure the data and annotation quality. We will explain them one by one.

\head{Challenges.} Here we list the challenging points in data collection and annotation, and provide our solutions and rationales behind. \textbf{(1) Source diverse videos of high-quality}. We show our principles in Sec.~\ref{sec:sub_video_collection}. \textbf{(2) How to ensure the quality of questions}: ask the ones people would naturally ask and balance the question types. We provide a verification-involved process in Sec.~\ref{sec:question_collection}.

\head{Summary of our AssistSR dataset.} In total, we collected 3,214 questions associated with 1,607 segments in 210 videos for the AssistSR dataset. We provide some examples of AssistSR along with qualitative results in Fig.~\ref{fig:prediction_examples}.

\subsection{Video Collection}
\label{sec:sub_video_collection}
We collect a set of videos which cover a wide range of scenarios and contain interesting and diverse contents. Here, each scenario refers to a category of commonly used item in daily life, such as vacuums, digital watches and so on.  


To source videos, we firstly maintain a scenario list which covers the majority of the highest level in HowTo100M~\cite{miech19howto100m} to ensure diversity. To obtain high-quality videos, we further set some criteria, which could be found in appendix. 
\begin{itemize}[leftmargin=*]
\setlength{\itemsep}{2pt}
\setlength{\parsep}{0pt}
\setlength{\parskip}{0pt}
    \item To utilize auto-generated ASR captions, we exclude videos without voice-over or with only scene-text.
    \item Videos should contain actions or salient motion for showcasing the feature of the subject item;
    \item Video contents should be rich enough to support addressing user’s question.
\end{itemize}

Overall, these videos are captured via different devices (e.g., mobile phone or GoPro) with different view angles (e.g., first-person view or third-person view), posing important challenges to computer vision systems. 
All the transcripts generated by ASR are in English. 

\begin{figure}[h]
\centering
\includegraphics[width=.9\linewidth]{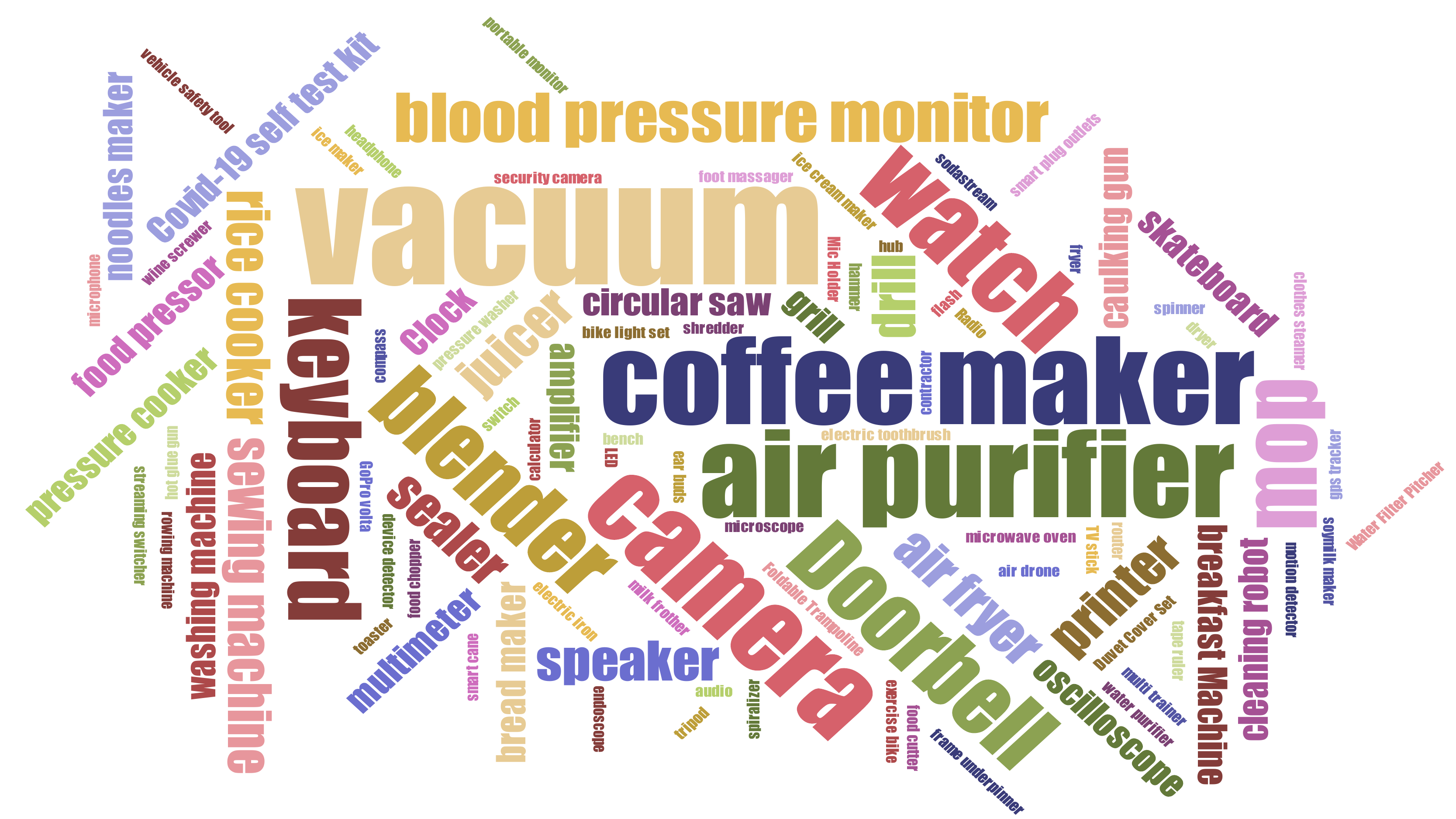}
\caption{High frequency scenarios in our dataset.}
\label{fig:scenario_word_cloud}
\end{figure}

In total, we collected 210 videos with an average duration of 539 seconds. All collected videos are mainly from the domain of commonly used items, including sub-categories ranging from home appliances, digital gadgets to smart devices. Fig.~\ref{fig:scenario_word_cloud} shows some scenarios with high frequency in our dataset.
In Tab.\ref{tab:cmp-assistant-dataset}, we compared AssistSR with some recently proposed datasets related to task-oriented operations or intelligent assistants: \textit{\#1. }TC-QA~\cite{tan2020task},  \textit{\#2. }TutorialVQA~\cite{colas-etal-2020-tutorialvqa} and \textit{\#3. }ScreenCast QA~\cite{ijcai2020-148tut}.
We can see that our AssistSR has comparable size to these datasets in terms of \#scenarios, \#videos, \#segments, \#QAs and total duration.
\begin{table}[h]\centering
\resizebox{0.99\columnwidth}{!}{%
\begin{tabular}{c|cccccc}
\hline
Datasets  & \#Scen.           & \#Vid. & \#Seg. & \#QAs & Tot. Duration  \\ \hline
\#1   & 1      & -       & 495       & 991   & 41.25 min          \\
\#2 & 1   & 76      & 408       & 6195  & -                 \\
\#3 & 1 & 76      & -         & 17768 & 333 min              \\
AssistSR     & 92            & 210     & 1607      & 3214  & 1887.9 min     \\ \hline
\end{tabular}
}
\caption{Statistics on various datasets related to topics of task-oriented operations or intelligent assistants.}\label{tab:cmp-assistant-dataset}
\end{table}

\subsection{Question Collection}
\label{sec:question_collection}


\head{Procedure for annotating questions.} To collect high-quality questions, the annotators were asked to: 
\begin{enumerate}[leftmargin=*]
\setlength{\itemsep}{2pt}
\setlength{\parsep}{0pt}
\setlength{\parskip}{0pt}
    \item \textit{Go through and understand the whole video}.
    \item \textit{Come up with a list of questions which can be clearly tackled after watching the video.} We exclude the questions that a user would not be interested in. For example, ``What color is this juicer?” which is superficial.
    \item \textit{Conduct self-verification and cross-verification on the question list.} Each annotator checks its question list and filters out the unqualified questions while for cross-verification, another annotator who has not watched the same video checks the questions from the view of a real user.
    \item \textit{Collect and annotate the qualified questions.} After harvesting the qualified textual questions, the annotators are required to collect and annotate the query images.
\end{enumerate}



Next, we provide details for our question collection and annotation.

\head{Multimodal query.} Each question annotated in AssistSR is multimodal, which means that it is composed of a textual part and a visual part. The textual part is a free-form query written by our annotators.
The visual part is an image for the object the annotator poses a question on. Inspired by ~\cite{mani2020point,zellers2019vcr}, we allow annotators to pose a question on specific parts of an object and locate its position with a bounding box. By incorporating the visual information, annotators are able to write more natural questions. For example, they can ask “How to clean up \colorbox{color_inp}{\texttt{<this part\includegraphics[scale=.07, bb=-50 25 150 14]{vacuum_bbox.png}>}}?” instead of “How to clean up the plastic dust bucket of my Roidme X20S vacuum?”,  with a bounding box tag
\colorbox{color_inp}{\texttt{<this part\includegraphics[scale=.07, bb=-50 25 150 14]{vacuum_bbox.png}>}}
referring to the specific part of the vacuum cleaner.

We also provide some strategies to avoid collecting query images from the target ground-truth video segment: searching different instances on the Internet, sampling a query image from non-GT segments with different contexts from GT segments, and editing images if they have to be sampled from GT segments. This makes TQVSR more difficult than simple visual matching. 
Details in appendix.


\head{Query type.} 
In practice, videos are associated with multi-modalities such as audio and text, e.g. subtitles for TV shows, transcripts for instructional videos or  audience discourse accompanying live stream, which could be also important sources for retrieving relevant segments. Following~\cite{lei2020tvr,lei2018tvqa}, we encourage annotators to write questions that are related to different modalities, aiming at enabling the model to learn the knowledge from the video in a systematic way. To test the model’s ability to retrieve the relevant segment for a given question, we categorize all the questions into 3 types: 
\begin{enumerate}[leftmargin=*]
\setlength{\itemsep}{0pt}
\setlength{\parsep}{0pt}
\setlength{\parskip}{0pt}
    \item \textbf{\textit{t type}}: the visual part of the annotated question only helps to match the subject in the video, one can retrieve the relevant clips by textual clues from videos besides such matching.
    \item \textbf{\textit{v type}}: one can only locate the relevant clips by visual clues from videos.
    \item \textbf{\textit{v+t type}}: one should locate the relevant segments by leveraging both visual and textual clues from videos. In this case, the visual part of the question not only helps to match the subject, but also provides vital information for segment retrieval.
\end{enumerate}

Similar to \cite{lei2020tvr, lei2018tvqa}, in our pilot test, we observed that our annotators preferred to write \textit{t type} questions, of which the answers segment can be easily retrieved by reading the transcripts. To ensure that we collect a balance of queries requiring one or both modalities, we set up awarding mechanism for asking \textit{v type} and \textit{t+v type} type questions. More examples of query types could be found in appendix.

\noindent\head{Overview of the collected questions.} We collected 3,214 questions in total. On average, our question contains 8.65 words. Each question is associated with one user image and with 0.31 bounding boxes annotated on average.  Fig.~\ref{fig:vid-query-length}b shows the distribution of question length, showing that most of the queries in our dataset have more than 10 words. 

\begin{figure}[t]
\centering
\includegraphics[width=\linewidth]{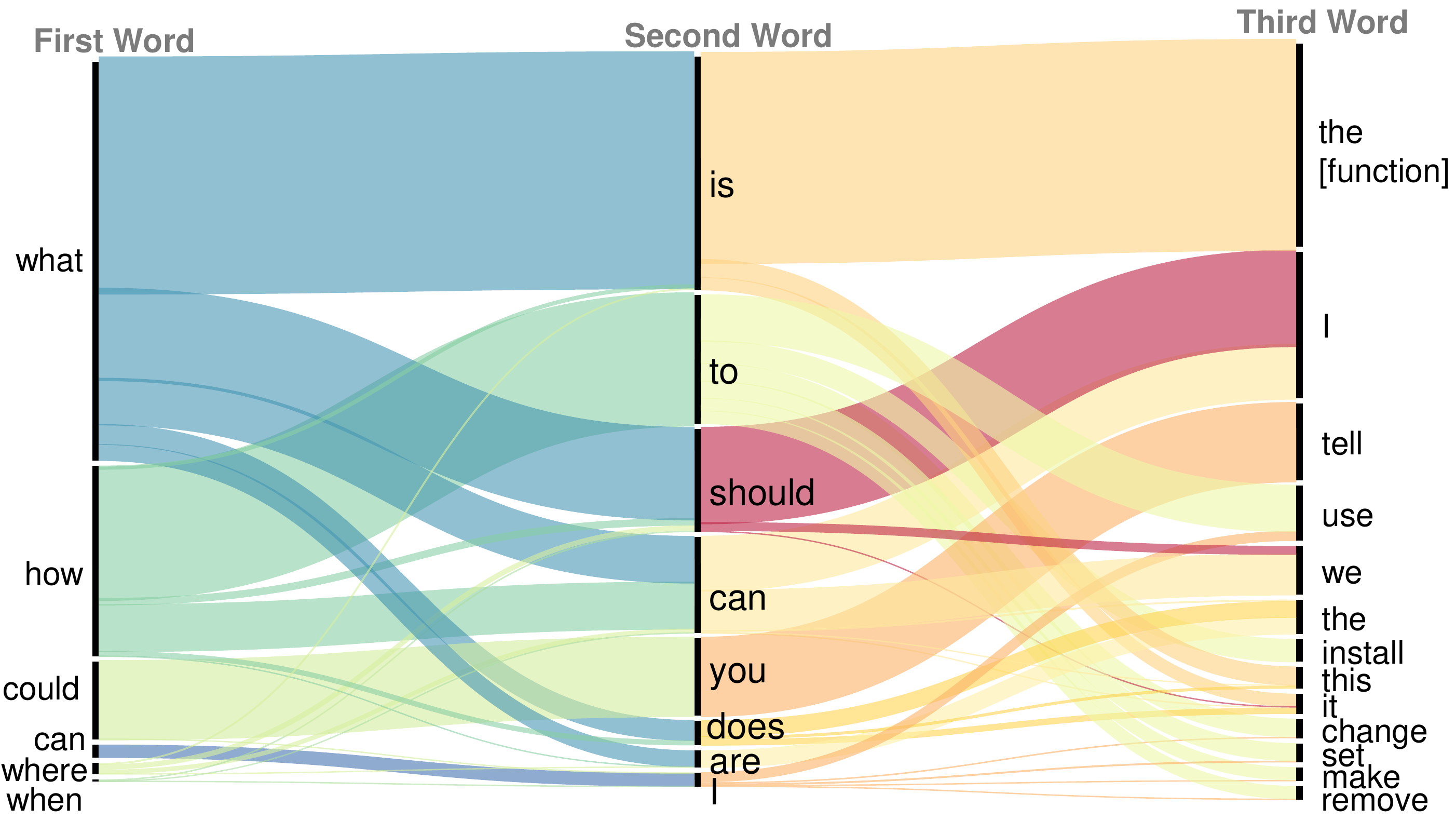}
\caption{
Alluvial diagram of the most frequent first three words in the collected questions.
}
\label{fig:dist_question_first_three}

\end{figure}

In Fig.~\ref{fig:dist_question_first_three}, 
by visualizing the Alluvial diagram of the most frequent first three words in the questions of our dataset, we observe that the text query pattern in our dataset is not limited to ``how to'' type questions. In real jobs, annotators can also ask questions with ``what'', ``when'', ``where'' and ``why''. Also, they are allowed to state the goal (e.g. ``I want to ...'') with a declarative sentence first and then ask a question. We can observe from Fig.~\ref{fig:dist_question_first_three} that ``what” type question is the most dominant in our dataset, followed by ``how to'' and ``could”.
The ``what'' type questions mainly ask about the functions of an item, and
the third word in ``how to” type questions are mainly verbs, showcasing the diversity and complexity of our questions, which is task-oriented.

\subsection{Video Segmentation}
Following Sec.~\ref{sec:video_seg}, we ask annotators to chunk a video into predefined segments and detailed analysis can be found in Sec.~\ref{sec:TQVSR_formulation} and appendix.

\begin{figure}[t]
  \centering
    \includegraphics[width=\linewidth]{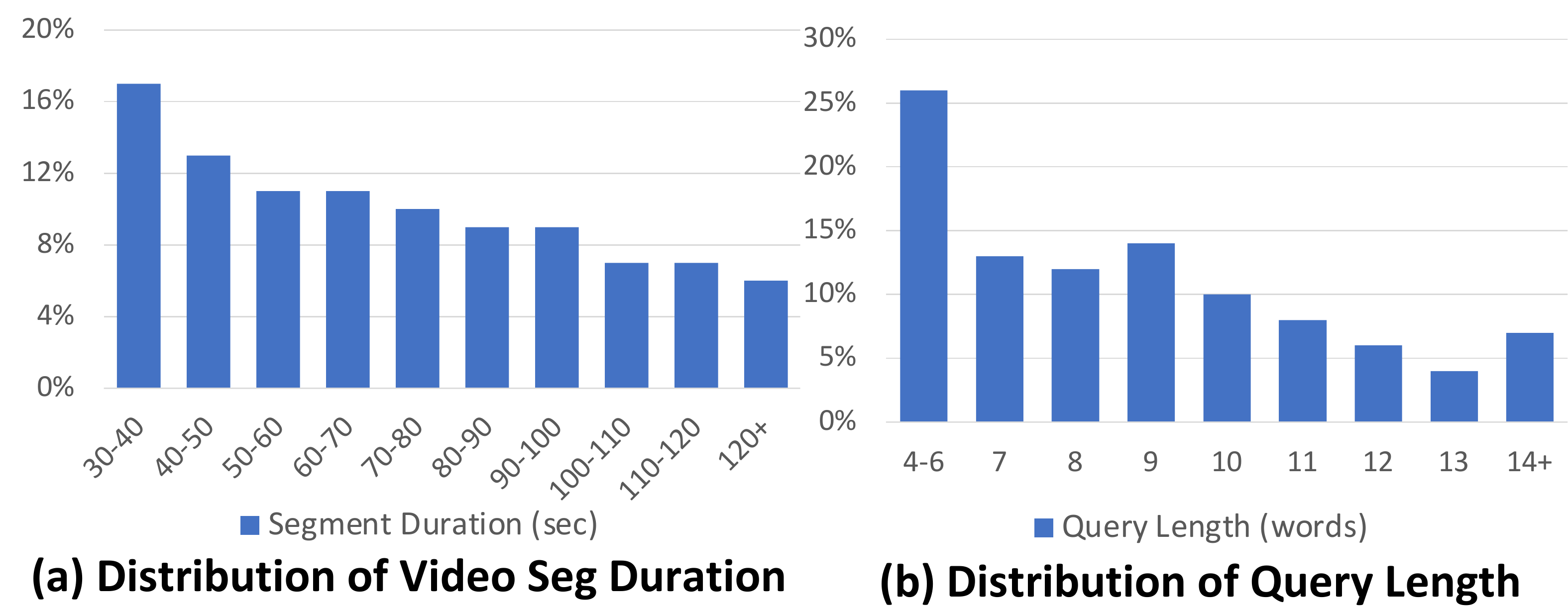}
  \caption{Distribution of segmentation and query length.}
  \label{fig:vid-query-length}
\end{figure}

Finally we collected 1,607 segments upon 210 videos.
The average duration for all chunked video segments is 70.5 seconds. Fig.~\ref{fig:vid-query-length}a shows the distribution of the length of the chunked video segments. Most of the video segments in our dataset have a duration longer than 50 seconds.

\subsection{Answer Annotation} 
With predefined segments, annotators are required to select the answer segments for each question. Each query with contents that can address the given question should be annotated. 
In AssistSR, there could be multiple disjoint answer segments paired with a single query (on average 1.03 answer segments per question), while all the moment localization or video retrieval datasets can only have one single moment or one single video. This is a more realistic setup as relevant content to a query in a video might be separated by irrelevant content.
\subsection{Quality Control} 
To ensure data and annotation quality, we’ve designed a detailed quality assurance mechanism, which is presented in appendix. Briefly, annotators were required to take the training curriculum and pass the qualification test before working on real jobs. Auditors are to evaluate the workers performance on the qualification test and check the quality of sampled annotations from the initial data pool. Typical issues in early training include asking unnatural questions, misunderstanding the rules for chunking a video, etc. Evaluation scores are rated on a scale of 1 (bad), 2 (minor errors) and 3 (good). Workers should retake the training curriculum when the rating score is deemed insufficient. In practice, the performance of an annotator is satisfying and acceptable if its average rating is above 2.5. More details could be found in appendix.

\section{Method}
\begin{figure*}[t]
  \centering
  \includegraphics[width=0.99\textwidth]{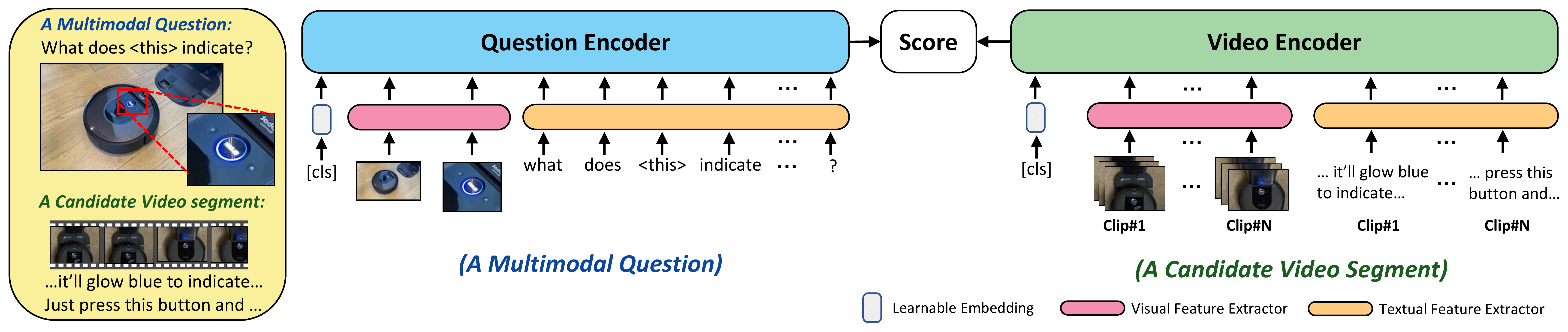}
  \caption{Illustration of our Dual Multimodal Encoders for TQVSR.}
  \label{fig:model}
\end{figure*}

We introduce \textbf{Dual Multimodal Encoders (DME)} for TQVSR. Fig.~\ref{fig:model} gives an overview of our model. Formally, we define the inputs into the model as: a candidate video segment associated with transcripts and a user question composed of an image with or without referring regions, and a textual query. 


\head{Model Architecture.} DME is composed of two multimodal encoders based on the transformer network~\cite{vaswani2017attention}. In TQVSR, both user question and a context video segment are multimodal, i.e. both of them are composed of vision input and language input, requiring a model for learning joint contextualized representations. Inspired by the recent success of the vision and language model ~\cite{Lu2019ViLBERTPT,Su2020VL-BERT:,li2019visualbert}, we follow~\cite{li2019visualbert} to use the self-attention mechanism within the Transformer to implicitly align elements of the input text and regions in the question and model the contextual information of the input video segment.

\head{Input Representation.} We extract visual features for video appearance, query image and region with ResNet-50 backbone~\cite{he2016deep} with weights from grid-feat~\cite{jiang2020gridfeat}, which is trained on Visual Genome~\cite{krishnavisualgenome}.
We extract contextualized text features using a 6-layer pretrained DistilBert~\cite{sanh2019distilbert}. Both visual features and textual features are projected into the same dimension, and added with token type embedding and positional embedding. We concatenate the text feature and visual feature to create a single sequence embedding and a learnable \texttt{[cls]} token~\cite{Devlin2019BERTPO} is concatenated to the beginning of the input feature, which is used to produce the final output representation of each transformer-based encoder.

\head{Training and Inference.} During training, one relevant segment is randomly sampled to be paired with a question. For training, we use loss function as in~\cite{zhai2018classification} for segment retrieval setting, where matching question-segment pairs in a batch are treated as positives, and all other pairwise combinations in the same batch are treated as negatives. 
We maximize the score between positive pairs and minimize the score between negative pairs.
At inference time, the DME model requires only the dot product between the multimodal question embedding and candidate video segment embeddings. 
This retrieval inference is of trivial cost since questions and video segments are indexable and therefore it is scalable to large scale retrieval.
More details are in appendix.

\section{Experiment}
\subsection{Baseline}
\label{subsec:intro_baseline}


\begin{figure*}[h]
\centering
\includegraphics[width=.99\linewidth]{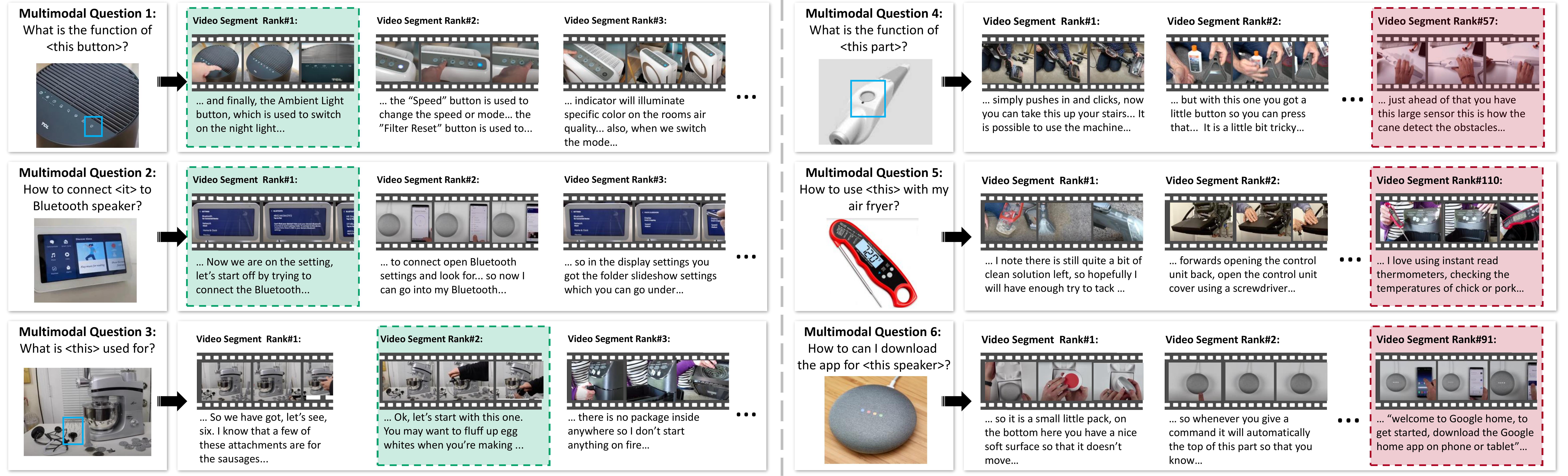}
\caption{DME prediction examples for TQVSR, for AssistSR dataset. Dashed box indicates the GT segments.
}
\label{fig:prediction_examples}
\end{figure*}

To explore the performance of other video-related tasks on TQVSR, 
we evaluate the following methods and modify the original implementation to fit in our inputs if necessary. More details could be found in appendix.

\head{\#1 Random guess.} For each question, we randomly shuffle the list of all video segments as the ranked results.
\textbf{\#2 SiameseNet}~\cite{chopra2005learning} for image-video matching.
\textbf{\#3 TVQA.}~\cite{lei2018tvqa} A multi-modal videoQA method in which QA pairs are used to fuse with visual and textual features in the video separately. 
We also replace LSTM~\cite{1997lstm} in the original implementation with transformer~\cite{vaswani2017attention}.
\textbf{\#4 XML}~\cite{lei2020tvr} is a late fusion approach for Video Corpus Moment Retrieval (VCMR). XML(VR) views each video segment in TQVSR as a video, and just compiles the Video Retrieval part in XML. XML(VR+ML) uses the rough start-end time in AssistSR annotation for VCMR setting. 
\textbf{\#5 ClipBERT}~\cite{lei2021less} is a video-language framework that sparsely and randomly sample video clips and frames within a clip for end-to-end training.

\subsection{Experimental Setting}
\head{Dataset.} We split our AssistSR into 80\% train, 10\% val and 10\% test such that videos and their associated queries appear in only one split. We include scenarios which are unseen in the train set into the validation set and test set. The distribution of query patterns and query types are kept aligned among these splits.

\head{Evaluation metrics.} To evaluate the performance of TQVSR, we use mean average precision (mAP) as in the information retrieval system. We also report standard metric Recall@1 (R@1) and Recall@5 (R@5) used in text-to-video retrieval and single moment retrieval. For TQVSR, we use mAP as the main metric. Note mAP is able to measure the model performance when the number of ground-truth video segments is greater than one.


\begin{table}[htb]
    \centering
    \resizebox{0.99\columnwidth}{!}{%
        \begin{tabular}{lcclcclcclcclcc}
        \toprule
        \multirow{2}{*}{Method} & \multicolumn{2}{c}{mAP}         &  & \multicolumn{2}{c}{R@1}         &  & \multicolumn{2}{c}{R@5}              \\ \cline{2-3} \cline{5-6} \cline{8-9}  
                                & All            & Unseen         &  & All            & Unseen        &  & All            & Unseen               \\ \midrule
        Random guess                   & 2.66          & 2.65          &  & 0.45           & 0.45          &  & 2.18          & 2.18                   \\
        SiameseNet                   & 11.77          & 11.50          &  & 3.90           & 3.42          &  & 14.19          & 13.68                   \\
        TVQA-LSTM                    & 13.30          & 11.51         &  & 1.24           & 0.69          &  & 23.51          & 20.04                   \\
        TVQA-Trans.                    & 14.46          & 12.99         &  & 5.63           & 3.59          &  & 19.52          & 18.06                   \\
        ClipBERT                & 18.14          & 17.67           &  & 7.43           & 6.92          &  & 27.23          & \textbf{26.92 }                   \\
        XML(VR)                 & 17.45          & 16.44           &  & 5.93           & 5.56          &  & 25.18          & 22.22         \\
        XML(VR + ML)            & 18.37          & 14.76          &  & 7.92           & 5.56          &  & 27.23          & 19.44                   \\
        DME(ours)                   & \textbf{22.92} & \textbf{20.44} &  & \textbf{11.94} & \textbf{9.61} &  & \textbf{30.13} & 25.62 \\ \hline
        DME(Unicoder) & 25.18 & 22.75 & & 13.51&10.65& & 34.61 & 33.45\\ \bottomrule
        \end{tabular}
        }
    \caption{Experimental results on various methods for TQVSR test set.}
    
    \label{tab:methods_cmp}
\end{table}
\subsection{Quantitative Results}
\noindent\head{Performances on all scenarios}. 
AssistSR val set results are shown in Tab.~\ref{tab:methods_cmp}. We can observe that:

\textbf{(1)} All the methods designed for video-related tasks are clearly better than random guess. DME clearly outperforms other methods for video-related tasks, indicating that models designed for other tasks can not generalize well to the TQVSR task. 

\textbf{(2)} SiameseNet for image-video matching performs worse than other baselines because it only considers visual matching, discarding the textual information in questions and transcripts. This also indicates that TQVSR is not a simple matching task, yet it requires the model to learn the textual knowledge behind.

\textbf{(3)} In TVQA, we see that TVQA-Trans. outperforms TVQA-LSTM, indicating that the transformer can better capture information from video segments. We also see that DME works better than TVQA, and this implies that the design of separate query-video appearance context matching and query-video transcript context matching may ignore to capture the alignment of video appearance and transcripts. 

\textbf{(4)} In ClipBERT, the sparsely sampled frames and their temporally aligned transcripts generated by ASR might fail to capture the global feature. As in HowTo100M~\cite{miech19howto100m}, it is likely that the transcript is not depicting the content in the sampled frame. This modeling design for multimodal video might fail to learn a good representation for global information while DME's fusion strategy for multimodal can avoid such a situation. 

\textbf{(5)} DME also outperforms both XML(VR) and XML(VR+ML). XML(VR+ML) with more supervision higher than XML(VR) in mAP. In XML, separate self-attention modules are used for different modalities in video retrieval. Its inferior performance compared to DME indicates that the multimodal encoder might be able to learn better representations for multimodal question and video representation, which leads to better performance on the TQVSR task.

Therefore, to tackle TQVSR, we need more customized designs for multi-modal video context modeling and query-video interaction.

\noindent\head{Performances on unseen scenarios}. From Tab.~\ref{tab:methods_cmp} We can observe that ClipBERT, XML and DME  generalize poorly to the unseen scenarios: the mAP score on the Unseen set is clearly lower than that on the All set. This is because for items within the same scenario could share similar knowledge, while for items for totally different scenarios, model may fail to bridge the gap of the difference in appearance, language style, etc.
Here we leave seeking better generalization on TQVSR for future work. All these confirm the challenging nature of TQVSR task and the need of our new benchmark.

\begin{figure}[ht]
    \centering
    \includegraphics[width=0.95\columnwidth]{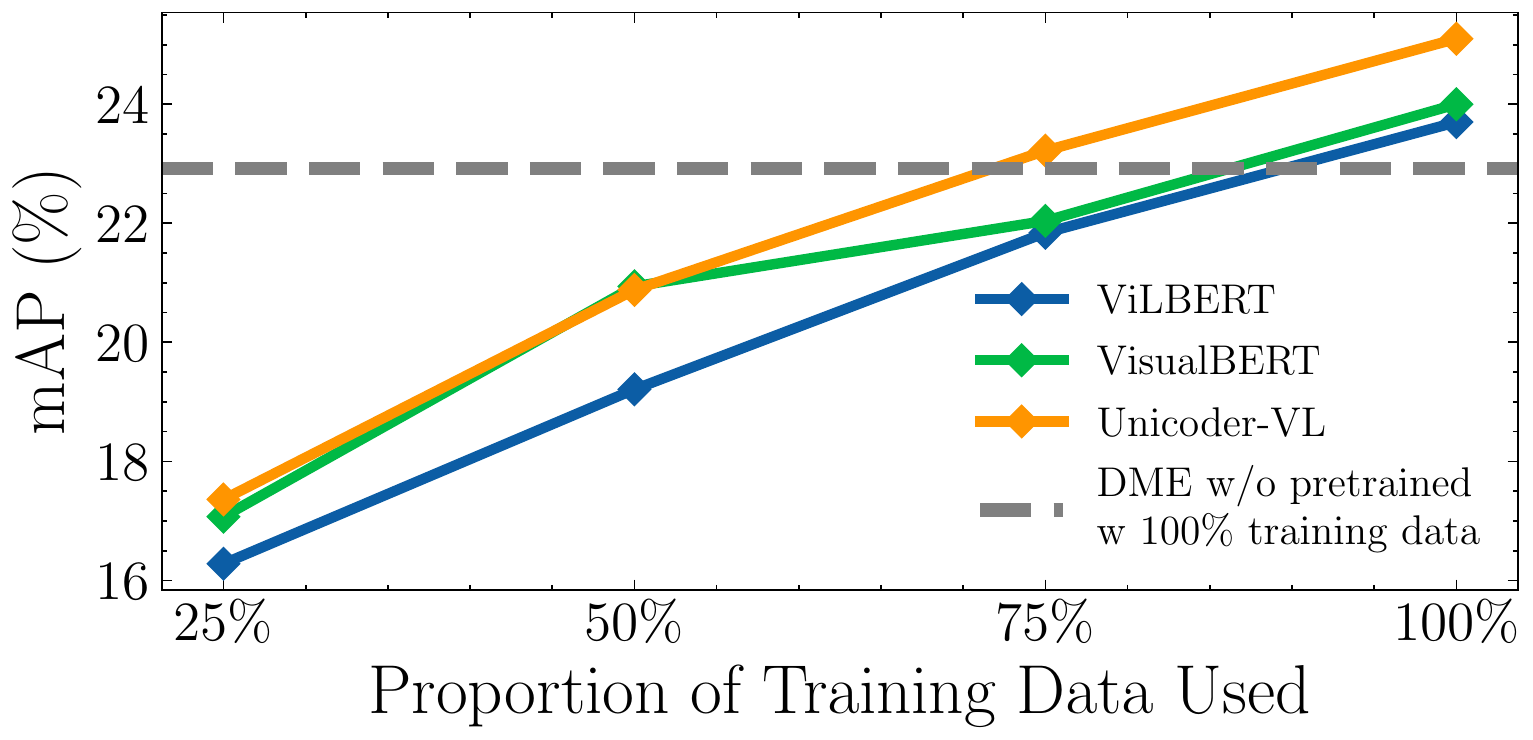}
    \captionof{figure}{Results of DME with pretrained weights on different proportion training data.}\label{fig:pretrained}
\end{figure}

\noindent\head{Would pretrained models help?}
For our proposed task, a promising direction is: leverage models pre-trained on other massive data and then only need a small amount of training data to finetune on our downstream task. 
We therefore initialized DME encoders from various pre-trained models: ViLBert~\cite{Lu2019ViLBERTPT}, VisualBert~\cite{li2019visualbert} and Unicoder-VL~\cite{li2020unicoder}, and then finetuned using 25\%, 50\%, 75\%, 100\% of AssistSR training set respectively. 
Fig.~\ref{fig:pretrained} shows that initializing DME encoders with various pre-trained models can improve DME (finetune w/ our whole train set). Also, when using only 75\% training data, using pre-trained model can achieve comparable mAP to DME with 100\% training data, indicating that proper vision and language model pretraining would help in the TQVSR task.

\begin{table*}[t]
\centering
\resizebox{.99\textwidth}{!}{
\begin{tabular}{lcccc|cccc|cccc|cccc}
\toprule[1pt]
             & \multicolumn{2}{c}{Question} & \multicolumn{2}{c|}{Video} & \multicolumn{4}{c|}{mAP}      & \multicolumn{4}{c|}{R@1}     & \multicolumn{4}{c}{R@5}       \\ \cline{6-17} 
             & Text          & Image        & Trans.       & Visual       & All   & T     & V     & T + V & All   & T     & V    & T + V & All   & T     & V     & T + V \\ \hline
\textit{\#1} & $\greencheck$       & $\greencheck$      & \redx            & $\greencheck$     & 13.36 & 12.94  & 12.8 & 14.08 & 4.5  & 3.95  & 9.62 & 3.1  & 17.42 & 16.45  & 9.62  & 21.71 \\
\textit{\#2} & \redx             & $\greencheck$      & $\greencheck$      & $\greencheck$     & 11.31 & 9.06 & 12.85  & 13.33 & 4.65  & 1.97  & 9.62 & 5.81  & 12.91 & 11.84 & 13.46  & 13.95 \\
\textit{\#3} & \redx             & $\greencheck$      & \redx            & $\greencheck$     & 10.29 & 8.61  & 12.08  & 11.56 & 3.00  & 0.66  & 7.69 & 3.88  & 14.11 & 14.47  & 9.62 & 15.50 \\ \hline
\textit{\#4} & $\greencheck$       & $\greencheck$      & $\greencheck$      & \redx           & 17.86 & 20.43 & 9.36 & 18.24 & 6.68  & 8.55  & 0.00 & 7.17 & 27.20 & 30.59 & 17.31  & 27.2 \\
\textit{\#5} & $\greencheck$       & \redx            & $\greencheck$      & $\greencheck$     & 18.87 & 25.65 & 9.21  & 14.79 & 8.36  & 12.50  & 2.88 & 5.68  & 26.25 & 38.16 & 11.54  & 18.15 \\
\textit{\#6} & $\greencheck$       & \redx            & $\greencheck$      & \redx           & 15.84 & 23.08 & 7.18  & 10.79 & 8.86  & 14.47  & 1.92 & 5.04  & 19.97 & 27.30 & 11.54  & 14.73 \\ \hline
\textit{\#7} & $\greencheck$       & $\greencheck$      & $\greencheck$      & $\greencheck$     & 22.92 & 27.37 & 18.29 & 19.54 & 11.94 & 13.82 & 11.54 & 9.88 & 30.13 & 40.13 & 19.23 & 22.74 \\ \bottomrule[1pt]
\end{tabular}
}
\caption{Ablation experiment results on various modalities for TQVSR.}
\label{tab:modal_ablition}
\end{table*}

\noindent\head{Modality removal from video source.} For the video encoder, we respectively mask the transcript (shown as Trans.) and the video appearance (shown as Visual). 
From \textit{\#1} and \textit{\#4} in Tab.~\ref{tab:modal_ablition},
we can see that mAP on \textit{All} drops significantly when either  transcript or video appearance is masked.
Performance drops more on \textit{T} and \textit{T+V} when masking transcript and leads to obvious degradation on \textit{V} when masking video appearance.

\noindent\head{What if removing modality from the user's question?}
For question encoder, we mask the textual and visual part of the query (shown as Text and Image in Tab.~\ref{tab:modal_ablition}) respectively and analyze the effect on the model performance. 
From \textit{\#2} and \textit{\#5}, we see that mAP on \textit{T+V} declines to almost the same when the textual or the visual part is masked. 
For {\textit{V}} questions, the mAP declines less when masking the textual part.
In contrast, the mAP on {\textit{T}} declines more when the text query is masked. 

\noindent\head{Textual vs. Visual.}
Comparing \textit{\#1-\#3} with \textit{\#4-\#6} on \textit{All} set, masking textual modality leads to a more drastic drop on mAP: this indicates that the textual modality is more important than visual modality. The reasons could be that: 
(1) the textual part especially the text query drives the retrieval task;
(2) DME over-relies on the textual information and lacks the alignment between text and vision, thus undermining the contribution of visual information. This can be observed from \textit{\#7}: with all modalities, the mAP of \textit{V} questions is lower than \textit{T} and \textit{T+V}.

Considering the complexity of our task, masking any modality could affect the model’s performance. Therefore, using multi-modalities is a direct way to improve the interaction quality.


\subsection{Qualitative Results}
We present qualitative examples in Fig.~\ref{fig:prediction_examples}. DME works well for many cases: for example, in the left column, it correctly retrieved the most relevant answer segment. Particularly, rank \#2 and rank \#3 results in \textit{col 1 row 2} show that it can retrieve segments with highly correlated contents. The right column shows the case where DME missed the correct segment.

\section{Conclusion and Future Work}
This paper presents a new task, Task-oriented Question-driven Video Segment Retrieval (TQVSR), requiring intelligent systems to address the given task-oriented question by retrieving relevant segments from a video corpus. To support the evaluation of such a task, we construct a dataset, AssistSR, which contains 3.2k questions on 1.6k video segments from instructional videos on diverse daily-used items. Further, we provide analyses of this new dataset as well as several baselines and a multi-stream end-to-end trainable neural network framework for TQVSR. The experimental results show that TQVSR is still a formidable challenge for all current methods. We hope this benchmark can help drive the vision and language models to be deployed in real-world applications and boost the development of personal AI Assistants.

\section*{Acknowledgement}
This project is supported by the National Research Foundation, Singapore under its NRFF Award NRF-NRFF13-2021-0008, and Mike Zheng Shou's Start-Up Grant from NUS. The computational work for this article was partially performed on resources of the National Supercomputing Centre, Singapore.

\section*{Limitations}
For the limitations of this work, we realize that the size of our dataset is limited due to the difficulties in the time consuming data collection and annotation process. We will make further efforts to create large-scale datasets in this topic, including widening for more scenarios and designing strategies to scale up data collection in a more efficient manner. 
For societal impact, we believe the proposed TQVSR is an important stepping stone towards building an intelligent agent. 
But to prevent the abuse of this technology, such as assisting unauthorized users to use dangerous items, we will take this into account and prohibit misuse cases explicitly in our license when releasing the developed codes and models.

\bibliography{anthology,custom}

\begin{thebibliography}{53}
\expandafter\ifx\csname natexlab\endcsname\relax\def\natexlab#1{#1}\fi

\bibitem[{how(2020)}]{howlong2020}
 2020.
\newblock How long should videos be? the ideal length for marketing.
\newblock [online].
\newblock
  \url{https://www.talkingtreecreative.com/blog/video-marketing-2/the-impact-of-video-length-on-engagement/}.

\bibitem[{vid(2020)}]{videolength2020}
 2020.
\newblock Video length: 4 tips that will help you boost engagement.
\newblock [online].
\newblock \url{https://meetmaestro.com/blogs/how-long-should-your-video-be/}.

\bibitem[{Anne~Hendricks et~al.(2017)Anne~Hendricks, Wang, Shechtman, Sivic,
  Darrell, and Russell}]{anne2017localizing}
Lisa Anne~Hendricks, Oliver Wang, Eli Shechtman, Josef Sivic, Trevor Darrell,
  and Bryan Russell. 2017.
\newblock Localizing moments in video with natural language.
\newblock In \emph{Proceedings of the IEEE international conference on computer
  vision}, pages 5803--5812.

\bibitem[{Antol et~al.(2015)Antol, Agrawal, Lu, Mitchell, Batra, Zitnick, and
  Parikh}]{antol2015vqa}
Stanislaw Antol, Aishwarya Agrawal, Jiasen Lu, Margaret Mitchell, Dhruv Batra,
  C~Lawrence Zitnick, and Devi Parikh. 2015.
\newblock Vqa: Visual question answering.
\newblock In \emph{Proceedings of the IEEE international conference on computer
  vision}, pages 2425--2433.

\bibitem[{Chopra et~al.(2005)Chopra, Hadsell, and LeCun}]{chopra2005learning}
Sumit Chopra, Raia Hadsell, and Yann LeCun. 2005.
\newblock Learning a similarity metric discriminatively, with application to
  face verification.
\newblock In \emph{2005 IEEE Computer Society Conference on Computer Vision and
  Pattern Recognition (CVPR'05)}, volume~1, pages 539--546. IEEE.

\bibitem[{Colas et~al.(2020)Colas, Kim, Dernoncourt, Gupte, Wang, and
  Kim}]{colas-etal-2020-tutorialvqa}
Anthony Colas, Seokhwan Kim, Franck Dernoncourt, Siddhesh Gupte, Zhe Wang, and
  Doo~Soon Kim. 2020.
\newblock \href {https://aclanthology.org/2020.lrec-1.670} {{T}utorial{VQA}:
  Question answering dataset for tutorial videos}.
\newblock In \emph{Proceedings of the 12th Language Resources and Evaluation
  Conference}, pages 5450--5455, Marseille, France. European Language Resources
  Association.

\bibitem[{Devlin et~al.(2019)Devlin, Chang, Lee, and
  Toutanova}]{Devlin2019BERTPO}
Jacob Devlin, Ming-Wei Chang, Kenton Lee, and Kristina Toutanova. 2019.
\newblock Bert: Pre-training of deep bidirectional transformers for language
  understanding.
\newblock In \emph{NAACL}.

\bibitem[{Gao et~al.(2017)Gao, Sun, Yang, and Nevatia}]{gao2017tall}
Jiyang Gao, Chen Sun, Zhenheng Yang, and Ram Nevatia. 2017.
\newblock Tall: Temporal activity localization via language query.
\newblock In \emph{Proceedings of the IEEE international conference on computer
  vision}, pages 5267--5275.

\bibitem[{Garcia~del Molino and Gygli(2018)}]{garcia2018phdgif}
Ana Garcia~del Molino and Michael Gygli. 2018.
\newblock \href {https://doi.org/10.1145/3240508.3240599} {Phd-gifs:
  Personalized highlight detection for automatic gif creation}.
\newblock pages 600--608.

\bibitem[{Gibson(1977)}]{gibson1977theory}
James~J Gibson. 1977.
\newblock The theory of affordances.
\newblock \emph{Hilldale, USA}, 1(2):67--82.

\bibitem[{Girshick(2015)}]{girshick2015fast-rcnn}
Ross Girshick. 2015.
\newblock Fast r-cnn.
\newblock In \emph{Proceedings of the IEEE international conference on computer
  vision}, pages 1440--1448.

\bibitem[{Girshick et~al.(2014)Girshick, Donahue, Darrell, and
  Malik}]{girshick2014rcnn}
Ross Girshick, Jeff Donahue, Trevor Darrell, and Jitendra Malik. 2014.
\newblock Rich feature hierarchies for accurate object detection and semantic
  segmentation.
\newblock In \emph{Proceedings of the IEEE conference on computer vision and
  pattern recognition}, pages 580--587.

\bibitem[{Gurari et~al.(2018)Gurari, Li, Stangl, Guo, Lin, Grauman, Luo, and
  Bigham}]{gurari2018vizwiz}
Danna Gurari, Qing Li, Abigale~J Stangl, Anhong Guo, Chi Lin, Kristen Grauman,
  Jiebo Luo, and Jeffrey~P Bigham. 2018.
\newblock Vizwiz grand challenge: Answering visual questions from blind people.
\newblock In \emph{Proceedings of the IEEE Conference on Computer Vision and
  Pattern Recognition}, pages 3608--3617.

\bibitem[{Gygli et~al.(2016)Gygli, Song, and Cao}]{micheal2016video2gif}
Michael Gygli, Yale Song, and Liangliang Cao. 2016.
\newblock \href {https://doi.org/10.1109/CVPR.2016.114} {Video2gif: Automatic
  generation of animated gifs from video}.
\newblock pages 1001--1009.

\bibitem[{He et~al.(2016)He, Zhang, Ren, and Sun}]{he2016deep}
Kaiming He, Xiangyu Zhang, Shaoqing Ren, and Jian Sun. 2016.
\newblock Deep residual learning for image recognition.
\newblock In \emph{Proceedings of the IEEE conference on computer vision and
  pattern recognition}, pages 770--778.

\bibitem[{Hochreiter and Schmidhuber(1997)}]{1997lstm}
Sepp Hochreiter and Jürgen Schmidhuber. 1997.
\newblock \href {https://doi.org/10.1162/neco.1997.9.8.1735} {Long short-term
  memory}.
\newblock \emph{Neural Computation}, 9(8):1735--1780.

\bibitem[{Jang et~al.(2017)Jang, Song, Yu, Kim, and Kim}]{jang2017tgif}
Yunseok Jang, Yale Song, Youngjae Yu, Youngjin Kim, and Gunhee Kim. 2017.
\newblock Tgif-qa: Toward spatio-temporal reasoning in visual question
  answering.
\newblock In \emph{Proceedings of the IEEE conference on computer vision and
  pattern recognition}, pages 2758--2766.

\bibitem[{Jiang et~al.(2020)Jiang, Misra, Rohrbach, Learned-Miller, and
  Chen}]{jiang2020gridfeat}
Huaizu Jiang, Ishan Misra, Marcus Rohrbach, Erik Learned-Miller, and Xinlei
  Chen. 2020.
\newblock In defense of grid features for visual question answering.
\newblock In \emph{Proceedings of the IEEE/CVF Conference on Computer Vision
  and Pattern Recognition}, pages 10267--10276.

\bibitem[{Kim et~al.(2017)Kim, Heo, Choi, and Zhang}]{kim2017deepstory}
Kyung-Min Kim, Min-Oh Heo, Seong-Ho Choi, and Byoung-Tak Zhang. 2017.
\newblock \href {https://doi.org/10.24963/ijcai.2017/280} {Deepstory: Video
  story qa by deep embedded memory networks}.
\newblock pages 2016--2022.

\bibitem[{Kingma and Ba(2015)}]{Kingma2015AdamAM}
Diederik~P. Kingma and Jimmy Ba. 2015.
\newblock Adam: A method for stochastic optimization.
\newblock \emph{CoRR}, abs/1412.6980.

\bibitem[{Krishna et~al.(2017)Krishna, Hata, Ren, Fei-Fei, and
  Carlos~Niebles}]{krishna2017dense}
Ranjay Krishna, Kenji Hata, Frederic Ren, Li~Fei-Fei, and Juan Carlos~Niebles.
  2017.
\newblock Dense-captioning events in videos.
\newblock In \emph{Proceedings of the IEEE international conference on computer
  vision}, pages 706--715.

\bibitem[{Krishna et~al.(2016)Krishna, Zhu, Groth, Johnson, Hata, Kravitz,
  Chen, Kalantidis, Li, Shamma, Bernstein, and Fei-Fei}]{krishnavisualgenome}
Ranjay Krishna, Yuke Zhu, Oliver Groth, Justin Johnson, Kenji Hata, Joshua
  Kravitz, Stephanie Chen, Yannis Kalantidis, Li-Jia Li, David~A Shamma,
  Michael Bernstein, and Li~Fei-Fei. 2016.
\newblock \href {https://arxiv.org/abs/1602.07332} {Visual genome: Connecting
  language and vision using crowdsourced dense image annotations}.

\bibitem[{Lei et~al.(2021{\natexlab{a}})Lei, Berg, and
  Bansal}]{lei2021qvhighlights}
Jie Lei, Tamara~L Berg, and Mohit Bansal. 2021{\natexlab{a}}.
\newblock Qvhighlights: Detecting moments and highlights in videos via natural
  language queries.
\newblock \emph{arXiv preprint arXiv:2107.09609}.

\bibitem[{Lei et~al.(2021{\natexlab{b}})Lei, Li, Zhou, Gan, Berg, Bansal, and
  Liu}]{lei2021less}
Jie Lei, Linjie Li, Luowei Zhou, Zhe Gan, Tamara~L. Berg, Mohit Bansal, and
  Jingjing Liu. 2021{\natexlab{b}}.
\newblock Less is more: Clipbert for video-and-language learningvia sparse
  sampling.
\newblock In \emph{CVPR}.

\bibitem[{Lei et~al.(2018)Lei, Yu, Bansal, and Berg}]{lei2018tvqa}
Jie Lei, Licheng Yu, Mohit Bansal, and Tamara~L Berg. 2018.
\newblock Tvqa: Localized, compositional video question answering.
\newblock In \emph{EMNLP}.

\bibitem[{Lei et~al.(2020{\natexlab{a}})Lei, Yu, Berg, and
  Bansal}]{lei2020tvqa+}
Jie Lei, Licheng Yu, Tamara Berg, and Mohit Bansal. 2020{\natexlab{a}}.
\newblock \href {https://doi.org/10.18653/v1/2020.acl-main.730} {Tvqa+:
  Spatio-temporal grounding for video question answering}.
\newblock pages 8211--8225.

\bibitem[{Lei et~al.(2020{\natexlab{b}})Lei, Yu, Berg, and Bansal}]{lei2020tvr}
Jie Lei, Licheng Yu, Tamara~L Berg, and Mohit Bansal. 2020{\natexlab{b}}.
\newblock Tvr: A large-scale dataset for video-subtitle moment retrieval.
\newblock In \emph{ECCV}.

\bibitem[{Li et~al.(2020)Li, Duan, Fang, Gong, and Jiang}]{li2020unicoder}
Gen Li, Nan Duan, Yuejian Fang, Ming Gong, and Daxin Jiang. 2020.
\newblock Unicoder-vl: A universal encoder for vision and language by
  cross-modal pre-training.
\newblock In \emph{Proceedings of the AAAI Conference on Artificial
  Intelligence}, volume~34, pages 11336--11344.

\bibitem[{Li et~al.(2019)Li, Yatskar, Yin, Hsieh, and Chang}]{li2019visualbert}
Liunian~Harold Li, Mark Yatskar, Da~Yin, Cho-Jui Hsieh, and Kai-Wei Chang.
  2019.
\newblock Visualbert: A simple and performant baseline for vision and language.
\newblock \emph{arXiv preprint arXiv:1908.03557}.

\bibitem[{Lu et~al.(2019)Lu, Batra, Parikh, and Lee}]{Lu2019ViLBERTPT}
Jiasen Lu, Dhruv Batra, Devi Parikh, and Stefan Lee. 2019.
\newblock Vilbert: Pretraining task-agnostic visiolinguistic representations
  for vision-and-language tasks.
\newblock In \emph{NeurIPS}.

\bibitem[{Maharaj et~al.(2017)Maharaj, Ballas, Rohrbach, Courville, and
  Pal}]{maharaj2017dataset}
Tegan Maharaj, Nicolas Ballas, Anna Rohrbach, Aaron Courville, and Christopher
  Pal. 2017.
\newblock A dataset and exploration of models for understanding video data
  through fill-in-the-blank question-answering.
\newblock In \emph{Proceedings of the IEEE Conference on Computer Vision and
  Pattern Recognition}, pages 6884--6893.

\bibitem[{Malle et~al.(2001)Malle, Moses, and Baldwin}]{malle2001intentions}
Bertram~F Malle, Louis~J Moses, and Dare~A Baldwin. 2001.
\newblock \emph{Intentions and intentionality: Foundations of social
  cognition}.
\newblock MIT press.

\bibitem[{Mani et~al.(2020)Mani, Yoo, Hinthorn, and
  Russakovsky}]{mani2020point}
Arjun Mani, Nobline Yoo, Will Hinthorn, and Olga Russakovsky. 2020.
\newblock \href {http://arxiv.org/abs/2011.13681} {Point and ask: Incorporating
  pointing into visual question answering}.

\bibitem[{Miech et~al.(2019)Miech, Zhukov, Alayrac, Tapaswi, Laptev, and
  Sivic}]{miech19howto100m}
Antoine Miech, Dimitri Zhukov, Jean-Baptiste Alayrac, Makarand Tapaswi, Ivan
  Laptev, and Josef Sivic. 2019.
\newblock How{T}o100{M}: {L}earning a {T}ext-{V}ideo {E}mbedding by {W}atching
  {H}undred {M}illion {N}arrated {V}ideo {C}lips.
\newblock In \emph{ICCV}.

\bibitem[{Mun et~al.(2017)Mun, Hongsuck~Seo, Jung, and Han}]{mun2017marioqa}
Jonghwan Mun, Paul Hongsuck~Seo, Ilchae Jung, and Bohyung Han. 2017.
\newblock Marioqa: Answering questions by watching gameplay videos.
\newblock In \emph{Proceedings of the IEEE International Conference on Computer
  Vision}, pages 2867--2875.

\bibitem[{Oates and Grayson(2004)}]{oates2004cognitive}
John~Ed Oates and Andrew~Ed Grayson. 2004.
\newblock \emph{Cognitive and language development in children.}
\newblock Open University Press.

\bibitem[{Over et~al.(2014)Over, Fiscus, Sanders, Joy, Michel, Awad, Smeaton,
  Kraaij, and Quénot}]{over2014trecvid}
Paul Over, Jon Fiscus, Gregory Sanders, David Joy, Martial Michel, George Awad,
  Alan Smeaton, Wessel Kraaij, and Georges Quénot. 2014.
\newblock Trecvid 2014 – an overview of the goals, tasks, data, evaluation
  mechanisms, and metrics.

\bibitem[{Regneri et~al.(2013)Regneri, Rohrbach, Wetzel, Thater, Schiele, and
  Pinkal}]{regneri2013grounding}
Michaela Regneri, Marcus Rohrbach, Dominikus Wetzel, Stefan Thater, Bernt
  Schiele, and Manfred Pinkal. 2013.
\newblock Grounding action descriptions in videos.
\newblock \emph{Transactions of the Association for Computational Linguistics},
  1:25--36.

\bibitem[{Rohrbach et~al.(2015)Rohrbach, Rohrbach, Tandon, and
  Schiele}]{rohrbach2015dataset}
Anna Rohrbach, Marcus Rohrbach, Niket Tandon, and Bernt Schiele. 2015.
\newblock A dataset for movie description.
\newblock In \emph{Proceedings of the IEEE conference on computer vision and
  pattern recognition}, pages 3202--3212.

\bibitem[{Sanh et~al.(2019)Sanh, Debut, Chaumond, and
  Wolf}]{sanh2019distilbert}
Victor Sanh, Lysandre Debut, Julien Chaumond, and Thomas Wolf. 2019.
\newblock Distilbert, a distilled version of bert: smaller, faster, cheaper and
  lighter.
\newblock \emph{arXiv preprint arXiv:1910.01108}.

\bibitem[{Song et~al.(2016)Song, Redi, Vallmitjana, and Jaimes}]{song2016click}
Yale Song, Miriam Redi, Jordi Vallmitjana, and Alejandro Jaimes. 2016.
\newblock \href {https://doi.org/10.1145/2983323.2983349} {To click or not to
  click: Automatic selection of beautiful thumbnails from videos}.
\newblock pages 659--668.

\bibitem[{Su et~al.(2020)Su, Zhu, Cao, Li, Lu, Wei, and Dai}]{Su2020VL-BERT:}
Weijie Su, Xizhou Zhu, Yue Cao, Bin Li, Lewei Lu, Furu Wei, and Jifeng Dai.
  2020.
\newblock \href {https://openreview.net/forum?id=SygXPaEYvH} {Vl-bert:
  Pre-training of generic visual-linguistic representations}.
\newblock In \emph{International Conference on Learning Representations}.

\bibitem[{Sun et~al.(2014)Sun, Farhadi, and Seitz}]{sun2014ranking}
Min Sun, Ali Farhadi, and Steve Seitz. 2014.
\newblock Ranking domain-specific highlights by analyzing edited videos.
\newblock In \emph{ECCV}.

\bibitem[{Tan et~al.(2020)Tan, Leong, Xu, Li, Fang, Cheng, Gauthier, Sun, and
  Lim}]{tan2020task}
Hui~Li Tan, Mei~Chee Leong, Qianli Xu, Liyuan Li, Fen Fang, Yi~Cheng, Nicolas
  Gauthier, Ying Sun, and Joo~Hwee Lim. 2020.
\newblock Task-oriented multi-modal question answering for collaborative
  applications.
\newblock In \emph{2020 IEEE International Conference on Image Processing
  (ICIP)}, pages 1426--1430. IEEE.

\bibitem[{Tapaswi et~al.(2016)Tapaswi, Zhu, Stiefelhagen, Torralba, Urtasun,
  and Fidler}]{tapaswi2016movieqa}
Makarand Tapaswi, Yukun Zhu, Rainer Stiefelhagen, Antonio Torralba, Raquel
  Urtasun, and Sanja Fidler. 2016.
\newblock Movieqa: Understanding stories in movies through question-answering.
\newblock In \emph{Proceedings of the IEEE conference on computer vision and
  pattern recognition}, pages 4631--4640.

\bibitem[{Vaswani et~al.(2017)Vaswani, Shazeer, Parmar, Uszkoreit, Jones,
  Gomez, Kaiser, and Polosukhin}]{vaswani2017attention}
Ashish Vaswani, Noam Shazeer, Niki Parmar, Jakob Uszkoreit, Llion Jones,
  Aidan~N Gomez, {\L}ukasz Kaiser, and Illia Polosukhin. 2017.
\newblock Attention is all you need.
\newblock \emph{Advances in neural information processing systems}, 30.

\bibitem[{Wolf et~al.(2020)Wolf, Debut, Sanh, Chaumond, Delangue, Moi, Cistac,
  Rault, Louf, Funtowicz, Davison, Shleifer, von Platen, Ma, Jernite, Plu, Xu,
  Scao, Gugger, Drame, Lhoest, and Rush}]{wolf-etal-2020-hugging-face}
Thomas Wolf, Lysandre Debut, Victor Sanh, Julien Chaumond, Clement Delangue,
  Anthony Moi, Pierric Cistac, Tim Rault, Rémi Louf, Morgan Funtowicz, Joe
  Davison, Sam Shleifer, Patrick von Platen, Clara Ma, Yacine Jernite, Julien
  Plu, Canwen Xu, Teven~Le Scao, Sylvain Gugger, Mariama Drame, Quentin Lhoest,
  and Alexander~M. Rush. 2020.
\newblock \href {https://www.aclweb.org/anthology/2020.emnlp-demos.6}
  {Transformers: State-of-the-art natural language processing}.
\newblock In \emph{Proceedings of the 2020 Conference on Empirical Methods in
  Natural Language Processing: System Demonstrations}, pages 38--45, Online.
  Association for Computational Linguistics.

\bibitem[{Xu et~al.(2016)Xu, Mei, Yao, and Rui}]{xu2016msr-vtt}
Jun Xu, Tao Mei, Ting Yao, and Yong Rui. 2016.
\newblock \href
  {https://www.microsoft.com/en-us/research/publication/msr-vtt-a-large-video-description-dataset-for-bridging-video-and-language/}
  {Msr-vtt: A large video description dataset for bridging video and language}.
\newblock IEEE International Conference on Computer Vision and Pattern
  Recognition (CVPR).

\bibitem[{Yu et~al.(2019)Yu, Xu, Yu, Yu, Zhao, Zhuang, and
  Tao}]{yu2019activitynetQA}
Zhou Yu, Dejing Xu, Jun Yu, Ting Yu, Zhou Zhao, Yueting Zhuang, and Dacheng
  Tao. 2019.
\newblock Activitynet-qa: A dataset for understanding complex web videos via
  question answering.
\newblock In \emph{Proceedings of the AAAI Conference on Artificial
  Intelligence}, volume~33, pages 9127--9134.

\bibitem[{Zellers et~al.(2019)Zellers, Bisk, Farhadi, and
  Choi}]{zellers2019vcr}
Rowan Zellers, Yonatan Bisk, Ali Farhadi, and Yejin Choi. 2019.
\newblock From recognition to cognition: Visual commonsense reasoning.
\newblock In \emph{The IEEE Conference on Computer Vision and Pattern
  Recognition (CVPR)}.

\bibitem[{Zhai and Wu(2018)}]{zhai2018classification}
Andrew Zhai and Hao-Yu Wu. 2018.
\newblock Classification is a strong baseline for deep metric learning.
\newblock \emph{arXiv preprint arXiv:1811.12649}.

\bibitem[{Zhao et~al.(2020)Zhao, Kim, Xu, and Jin}]{ijcai2020-148tut}
Wentian Zhao, Seokhwan Kim, Ning Xu, and Hailin Jin. 2020.
\newblock Video question answering on screencast tutorials.
\newblock In \emph{Proceedings of the Twenty-Ninth International Joint
  Conference on Artificial Intelligence, {IJCAI-20}}, pages 1061--1068.
  International Joint Conferences on Artificial Intelligence Organization.
\newblock Main track.

\bibitem[{Zhu et~al.(2017)Zhu, Xu, Yang, and Hauptmann}]{zhu2017uncovering}
Linchao Zhu, Zhongwen Xu, Yi~Yang, and Alexander~G Hauptmann. 2017.
\newblock Uncovering the temporal context for video question answering.
\newblock \emph{International Journal of Computer Vision}, 124(3):409--421.

\end{thebibliography}
\bibliographystyle{acl_natbib}

\appendix
\section*{Appendix}
\noindent\head{- List of Contents in Appendix:}

\noindent\head{A. More Details of AssistSR Dataset.}

\noindent\head{B. More Details on Dataset Creation.}

\noindent\head{C. More Details of DME Method.}

\noindent\head{D. More Details of Experiment.}

\begin{figure*}[htp]
\centering
\includegraphics[width=.99\linewidth]{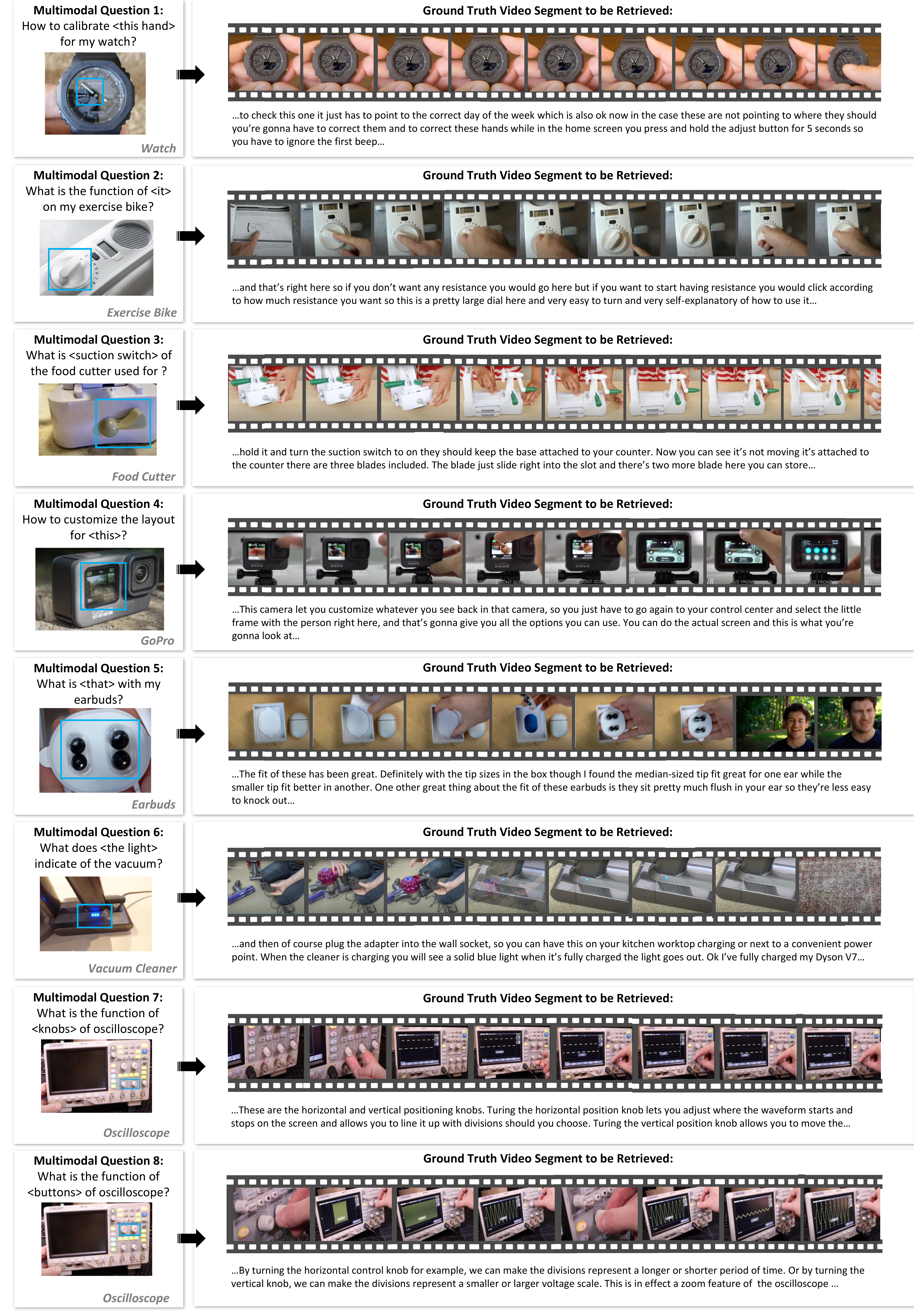}
\caption{Examples in the AssistSR dataset.}
\label{fig:supp_examples}

\end{figure*}

\section{More details of AssistSR Dataset}
\label{sec:supp_dataset}
\begin{table*}[h]
\centering
\resizebox{.9\textwidth}{!}{
\begin{tabular}{ccccccc}
\toprule[1pt]
\multirow{2}{*}{Dataset} & \multirow{2}{*}{Video source} & \multicolumn{2}{c}{Query source} & \multicolumn{2}{c}{Query modal} & \multirow{2}{*}{Affordance-centric} \\
                         &                         & Textual             & Visual             & Textual        & Visual         &                                   \\ \hline
Movie-QA~\cite{tapaswi2016movieqa}                 & movie                   & $\greencheck$       & -             & $\greencheck$        & -              & -                                 \\
TVQA~\cite{lei2018tvqa}                     & TV show                 & $\greencheck$       & $\greencheck$       & $\greencheck$        & -              & -                                 \\
TACoS~\cite{regneri2013grounding}                    & cooking                 & $\greencheck$       & -             & $\greencheck$        & -              & -                                 \\
DiDeMo~\cite{anne2017localizing}                   & Flickr                  & $\greencheck$       & -             & $\greencheck$        & -              & -                                 \\
ActivityNet Cap~\cite{krishna2017dense}         & Activity                & $\greencheck$       & -             & $\greencheck$        & -              & -                                 \\
TVR~\cite{lei2020tvr}                      & TV show                 & $\greencheck$       & $\greencheck$       & $\greencheck$        & -              & -                                 \\
ScreenCast QA~\cite{ijcai2020-148tut}            & Software tutorial       & $\greencheck$       & $\greencheck$       & $\greencheck$        & -              & $\greencheck$                           \\
ours                     & Tutorial for devices    & $\greencheck$       & $\greencheck$       & $\greencheck$        & $\greencheck$        & $\greencheck$                           \\ \bottomrule[1pt]
\end{tabular}
}
\caption{Comparison of AssistSR to various existing VideoQA/Video Retirieval/Moment Localization datasets.}
\label{tab:dataset_cmp}
\end{table*}

\subsection{Comparison with other datasets.} In Tab.~\ref{tab:dataset_cmp}, we compare AssistSR to previous VideoQA, Video Retrieval, Video Moment Localization datasets. In summary, our dataset is possessed with multimodal questions. 
In addition, we provide affordance-centric questions, aiming to drive the models to learn the knowledge from multimodal instructional videos to address the user problem beyond simple facts.
Besides, AssistSR can have multiple disjoint answer segments paired with a single query (on average 1.03 answer segments per question), while all the moment localization or video retrieval datasets can only have one single moment or one single video. This is a more realistic setup as relevant content to a query in a video might be separated by irrelevant content.

\subsection{Visualization Examples}
In Fig.~\ref{fig:supp_examples}, we show some visualization examples in our datasets. Specifically, we show examples of different scenarios (different devices as shown in row 1 - row 6 in Fig.~\ref{fig:supp_examples}) and examples of different functions of one scenario (as shown in row 7 - row 8 in Fig.~\ref{fig:supp_examples}).

In AssistSR, each multimodal question is associated with an image (with or without bounding boxes for referring to a specific region) and a human-written query. GT video segments contain the content which can address the given question.

\section{More Details on Dataset Creation}
\label{sec:supp_dataset_creation}
\subsection{Video Collection}
For video collection, we focus on commonly used items which can support asking affordance-centric questions. To improve efficiency for sourcing video, we provide some \textbf{tips for sourcing pattern}:
\begin{itemize}[leftmargin=12pt]
\setlength{\itemsep}{2pt}
\setlength{\parsep}{0pt}
\setlength{\parskip}{0pt}
    \item ``Tutorials/Instructional videos for ..."
    \item ``Best tips \& tricks for ..."
    \item ``How to use a ..."
    \item Searching by the name of a brand, e.g. Xiaomi.
    \item Check the recommendation list.
    \item Check the Youtuber's channel if you found several videos of his/hers are qualified.
\end{itemize}

We also set some criteria for quality assurance:
\begin{itemize}[leftmargin=*]
\setlength{\itemsep}{2pt}
\setlength{\parsep}{0pt}
\setlength{\parskip}{0pt}
    \item To utilize auto-generated ASR captions, we exclude videos without voice-over or with only scene-text.
    \item Videos should contain actions or salient motion for showcasing the feature of the subject item;
    \item Video contents should be rich enough to support addressing user’s question.
\end{itemize}

\subsection{Question Collection}

\begin{figure}[H]
\centering
\includegraphics[width=.9\linewidth]{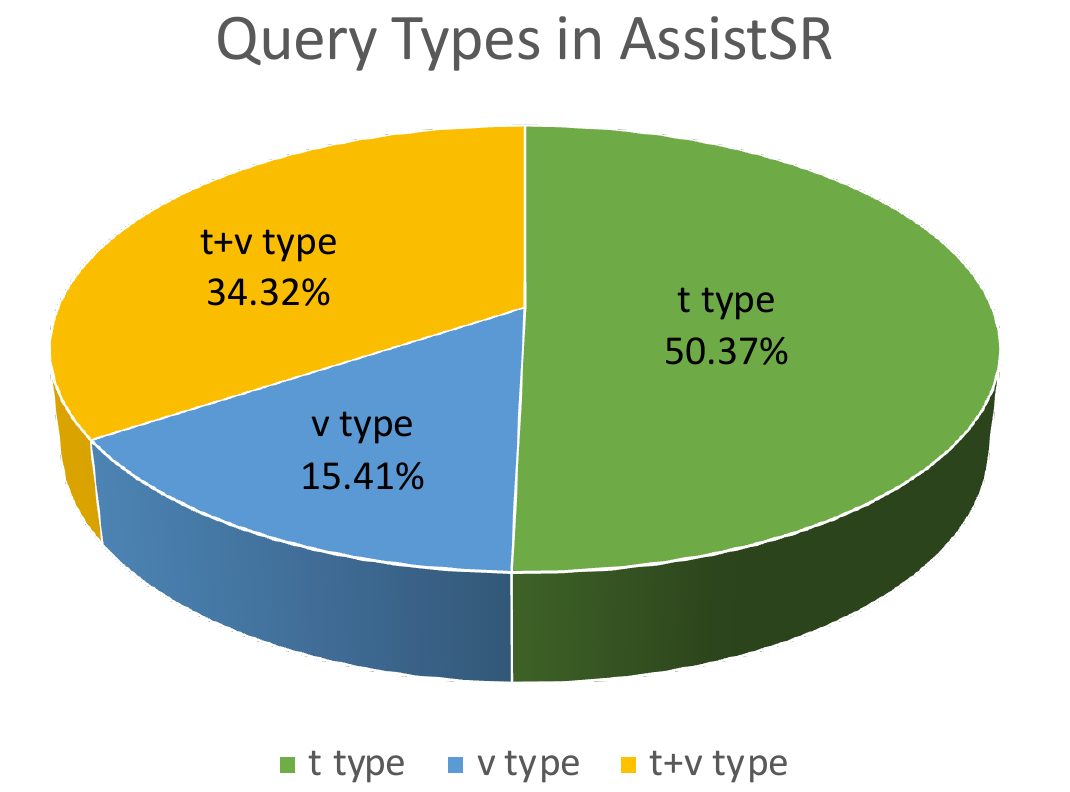}
\caption{Query type distribution of the AssistSR dataset.}
\label{fig:supp_query_types_all}
\end{figure}

\head{Query type.} To test the model’s ability to retrieve the relevant segment for a given question, we categorize all the questions into 3 types: \textbf{\textit{v type}}, \textbf{\textit{t type}} and \textbf{\textit{t+v type}}.

\begin{figure*}[htp]
\centering
\includegraphics[width=.99\linewidth]{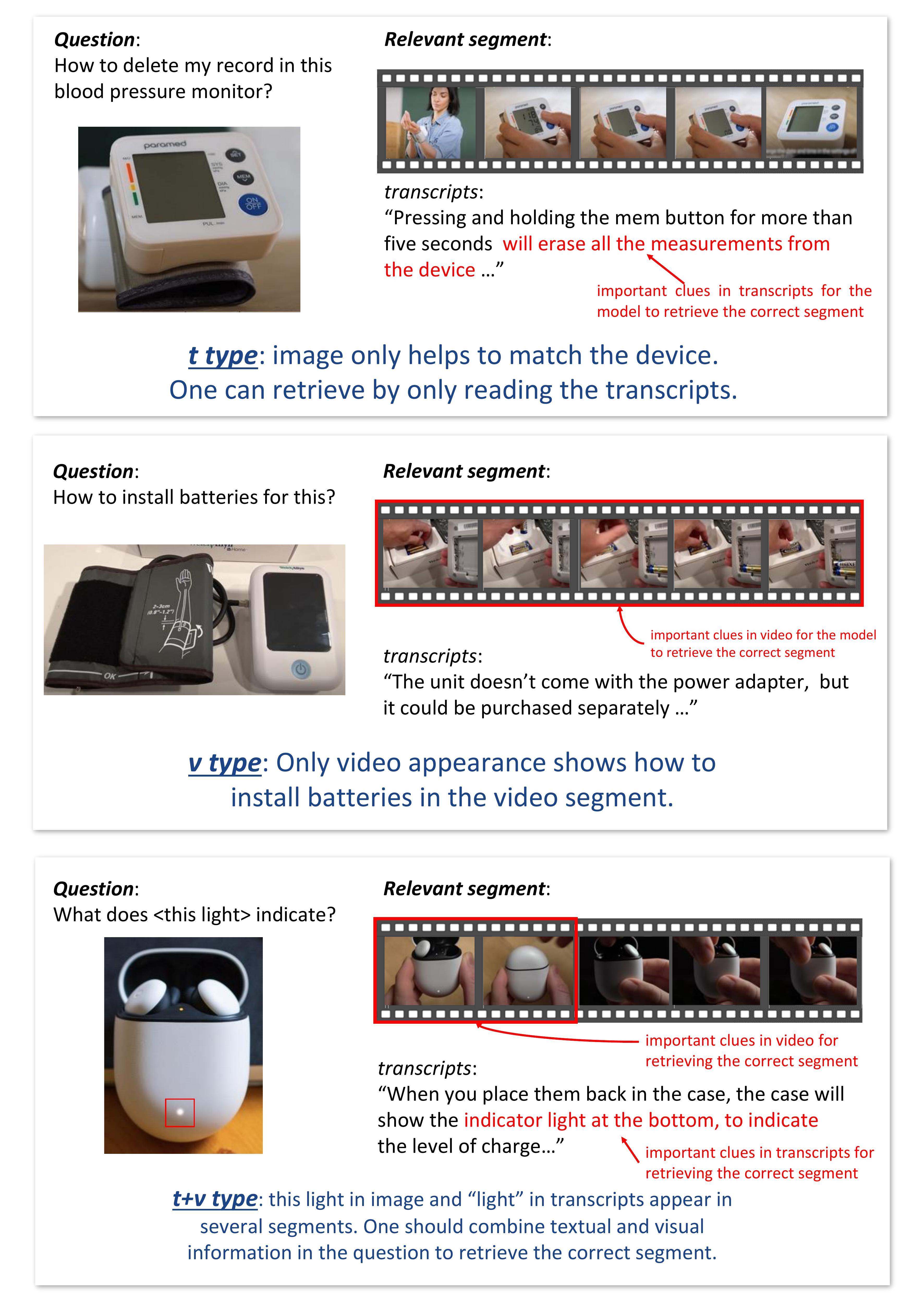}
\caption{Examples of different query types in the AssistSR dataset.}
\label{fig:supp_query_type_examples}
\end{figure*}

Similar to \cite{lei2020tvr, lei2018tvqa}, in our pilot test, we observed that our annotators preferred to write \textit{t type} questions, of which the answers segment can be easily retrieved by reading the transcripts. 
To ensure that we collect a balanced set of queries requiring one or both modalities, we set up awarding mechanism for asking \textit{v type} and \textit{t+v type} questions. In Fig.~\ref{fig:supp_query_type_examples} we provide examples of different query types and also explain why it belongs to that specific type.
We also provide these examples in Supp video.

In Fig.~\ref{fig:supp_query_types_all}, we show the distribution of the query types for our collected questions in the AssistSR dataset. With the help of the awarding mechanism, we balanced the number of \textit{t type} and not-\textit{t type} questions in our dataset.

\head{Multimodal query.} Human psychology literature shows that pointing to interesting objects or situations is one of the first ways by which babies communicate intention \cite{mani2020point, oates2004cognitive,malle2001intentions}.  The multimodal query adopted in AssistSR not only helps to provide an unambiguous link between the textual description of an object and the corresponding image region, but also avoids unnatural questions in pure text format. 

\begin{figure}[H]
\centering
\includegraphics[width=.9\linewidth]{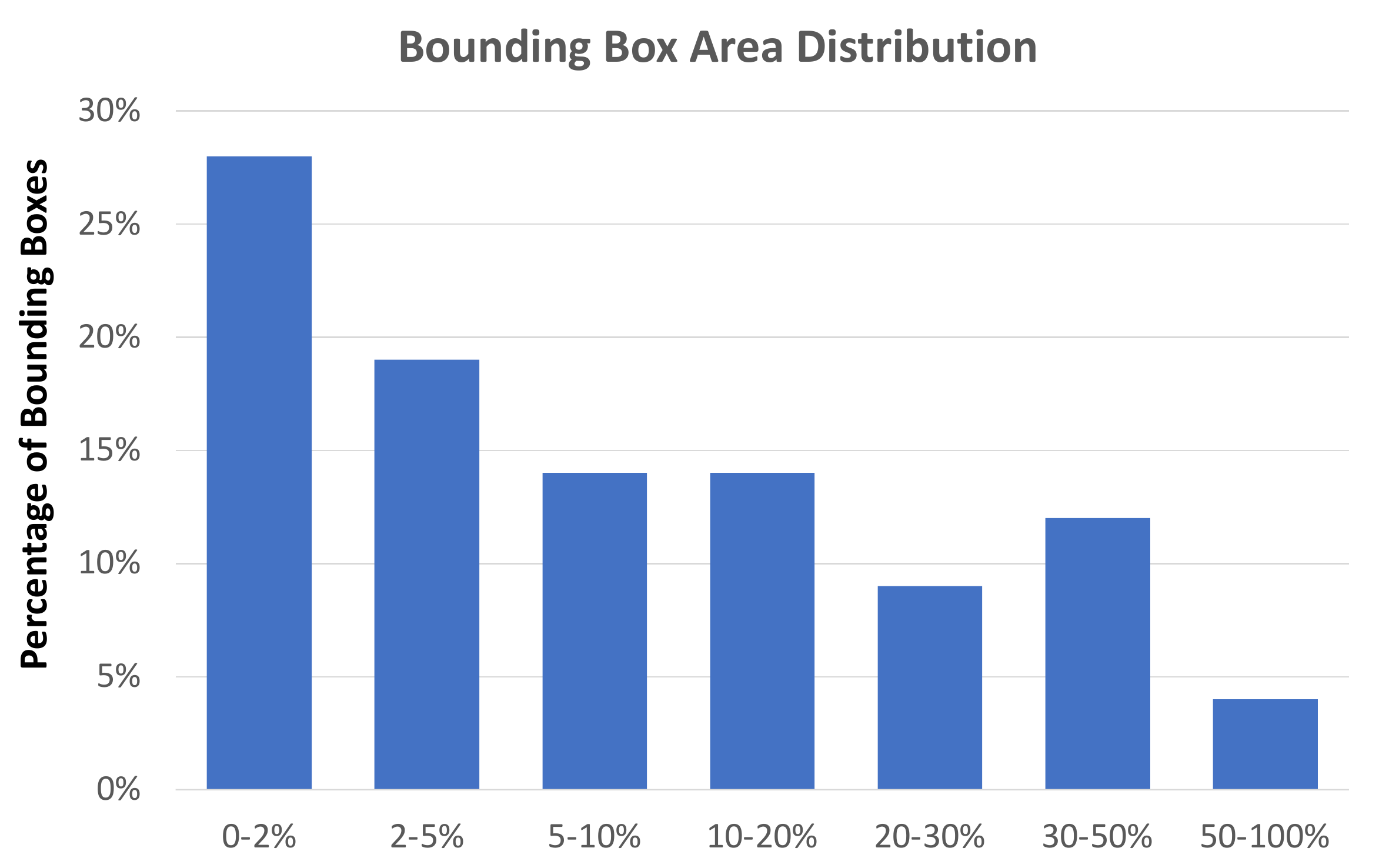}
\caption{Distribution for $\rm \frac{Area(bbox)}{Area(img)}$.}
\label{fig:supp_bbox_area}
\end{figure}

In AssistSR, the annotators are required to provide an image from a user's view, indicating the subject one would pose a question on. They are allowed to annotate bounding boxes for a specific part if it is needed in asking a question. Overall, each collected question is associated with an image, and on average 0.31 bounding boxes. In Fig.~\ref{fig:supp_bbox_area}, we show the distribution of the size of bounding boxes, in terms of the bounding box area over the whole image.

\head{Source of User Image.} Most query images are NOT from the Ground-Truth (GT) segments to be retrieved. To do so,
\begin{enumerate}[leftmargin=*]
\setlength{\itemsep}{2pt}
\setlength{\parsep}{0pt}
\setlength{\parskip}{0pt}
    \item Annotators are asked to first search on Internet to find images of the same device but different instance.
    \item If not successful, annotators will first source query image from non-GT segments of different contexts.
    \item If still not successful, annotators have to crop query image from GT segments, but we ask them to edit the original images (e.g. jiterring, flipping, brightness).
\end{enumerate}

As a result, 69.15\% of the query images are sourced from the internet and 8.28\% are from a non-ground-truth segment. All rows in Fig.~\ref{fig:prediction_examples} the 2nd column show such examples. Thus, most query images in AssistSR are of different instances and/or different contexts.

\subsection{Video Segmentation}
\head{Principles and rationales for video segmentation.} We provide principles for video segmentation in Sec. 3 in our paper. Here we provide more details: give a video, we ask the annotator to segment the videos following these principles:
\begin{enumerate}[leftmargin=12pt]
\setlength{\itemsep}{2pt}
\setlength{\parsep}{0pt}
\setlength{\parskip}{0pt}
    \item \textit{Keep consistent level of semantic granularity within one video segment.} The idea is that we want the content within one video segment to be semantically consistent. For example, a qualified segment could be the one introducing different aspects of a subject or introducing partial contents of an aspect (i.e. we can divide the content of this aspect into several segments).
    \item \textit{Soft threshold for video segment duration: 30 seconds as minimum and 2 minutes as maximum.} User study~\cite{howlong2020,videolength2020}, found that video clips should fully engage the audience within the first 30 seconds to attract their attention. Also, it is observed that keeping videos shorter than 2 minutes achieves the most viewer engagement while engagement drops off sharply after 2 minutes. Therefore, in video segmentation, we restrict the duration of each segment by setting a soft threshold.
    \item \textit{Adjust segmentation according to the list of qualified questions.} Since we adopt a flexible manner for segmenting a video, the result segmentation is not unique and thus adjustable. In order to test the models ability on retrieving the correct answer segment, we further ask our annotators to adjust the segmentation based on the raised question. For example, if one video segment contains both contents of correct answer and distraction of a question, the annotator should separate this into a positive segment and a hard negative segment.
\end{enumerate}
We provide examples for video segmentation in Supp video.

\subsection{Quality Control}
\label{sec:quality_control}
Here we provide the detailed quality assurance guideline for the auditor's evaluation. As shown in our main paper,  auditors are to evaluate the workers performance on the qualification test and check the quality of the sampled annotations from the initial data pool. 
To ensure the annotation quality, 
(1) Before working on real jobs, the annotator should carefully go through our guideline and annotate 5 videos (video collection, question collection, video segmentation, answer annotation) and auditors will audit annotators’ performance. If the performance is not satisfying, the annotator should retake the training curriculum.
(2) During the real annotation process, we conduct auditing for each annotator every iteration by randomly inspecting samples they completed in that iteration.

In the following, we firstly introduce the rating score definition for auditing, followed by details for different sub-tasks in annotation.

\subsubsection{Rating score definition} \textbf{1, bad}: The sourced data / annotation is of low quality. E.g, the annotator sourced unqualified videos for subsequent annotations; asked unnatural/uninteresting questions too often, chunked the videos mistakenly; wrong query type.
\textbf{2, medium}: occasional error. E.g less than 20\% questions are unnatural/uninteresting, but can be improved. The main purpose of this intermediate level is to reflect the need of improvement for the annotator while acknowledging the annotator’s correct understanding of the guideline.
\textbf{3, good}: source high-quality videos. Good questions, segmentations, etc.

\subsubsection{Video collection} \textbf{1, bad}: videos without voice (thus cannot obtain transcripts by ASR); videos without salient action(only scene text; only introduce the features by words); videos cannot support any interesting questions; more than 20\% of videos do not meet the requirements.
\textbf{2, medium}: 10\%-20\% videos do not meet the requirements in the guideline.
\textbf{3, good}: satisfies the requirements in our guideline: videos with voice (someone is explaining how to use the device/ finish a task; transcripts can be automatically generated); videos with salient action (someone is showing how to use the item); videos can support asking affordance-centric questions.

\subsubsection{Question collection} We firstly define unnatural questions and uninteresting questions with examples. Unnatural questions refer to the ones a user would not naturally ask in daily life. For example, for a knob on a microwave, a user might not ask “What would happen if I turn this clockwise?”. Instead, he would ask “How to set the heating time to 90 seconds” for using purpose. Uninteresting questions refer to the ones which are easy, e.g. “How to turn it on?”, the answer of which is obvious.
\textbf{1, bad}: 
more than 20\% questions are unnatural or uninteresting; 
miss \textit{v type} questions if it is obvious that one video is found to support asking several questions of this type; 
mistakes in annotating query type: the correct type should be \textit{t type} but annotated as \textit{t+v type}; the correct type should be \textit{t+v type} / \textit{v type} but annotated as \textit{t type}, the correct type should be \textit{t type} but annotated as \textit{v type}, the correct type is \textit{v type} but annotated as \textit{t type}.
\textbf{2, medium}: 
occasionally ask unnatural/uninteresting questions, but the questions can be rephrased to be good ones;
occasional minor mistakes in annotating query type: the correct type should be \textit{t+v type} but annotated as \textit{v type} , the correct type should be \textit{v type} but annotated as \textit{t+v type}.
\textbf{3, good}: 
Annotation satisfies the requirements in our guideline: ask natural and interesting questions, the textual and visual parts of the query make the question clear and easy to understand; understand the meaning of the query type and annotate correctly.

\subsubsection{Video segmentation} 
\textbf{1, bad}: non-consistent semantic level within one segment; there exists short segments (obviously less than 30s) which should be merged with other segments; do not refine the segmentation based on the questions.
\textbf{2, medium}: occasionally missed the segmentation adjusting the segmentation.
\textbf{3, good}; follow the three principles for video segmentation.

\subsubsection{Answer annotation}
\textbf{1, bad}: miss important contents to answer that question; mistakenly include totally irrelevant segments.
\textbf{3, good}: the annotator correctly annotates all the relevant segments.

\subsection{Annotator Recruitment}
For dataset creation, we hired annotators from a university. We set up a training curriculum and hire those who pass the pilot test (criteria in \ref{sec:quality_control}). Finally 8 annotators are proceeded to the real job for data collection and annotation.
The dataset creation costs around 600 man hours, with 10 USD/(man hour).

\begin{table*}[h]
 \centering
 \resizebox{.99\textwidth}{!}{
 \begin{tabular}{lclclclclclclclclclc}
\toprule[1pt]
\multirow{2}{*}{Method} & \multicolumn{3}{c}{mAP}                                                            &  & \multicolumn{3}{c}{R@1}                                                           &  & \multicolumn{3}{c}{R@5}                                                            &  & \multicolumn{3}{c}{R@10}                                                           &  & \multicolumn{3}{c}{R@50}                                                           \\ \cline{2-4} \cline{6-8} \cline{10-12} \cline{14-16} \cline{18-20} 
                                                      & All                             & Seen           & Unseen                          &  & All                             & Seen           & Unseen                         &  & All                             & Seen           & Unseen                          &  & All                             & Seen           & Unseen                          &  & All                             & Seen           & Unseen                          \\ \hline
TVQA~\cite{lei2018tvqa}         & 13.30                           & 14.29          & 11.51                           &  & 1.24                            & 1.54           & 0.69                           &  & 23.51                           & 25.38          & 20.14                           &  & 35.15                           & 36.15          & 33.33                           &  & 78.96                           & 83.46          & 70.83                           \\
ClipBERT~\cite{lei2021less}     & 18.14                           & 19.00          & 17.67                            &  & 7.43                            & 8.33           & 6.92                           &  & 27.23                           & 27.78          & \textbf{26.92}                            &  & 41.34                           & 38.89          & 42.69                           &  & 80.69                           & 75.00          & \textbf{83.85}                           \\
XML(VR)~\cite{lei2020tvr}       & 17.45                           & 17.99          & 16.44                            &  & 5.93                            & 6.13           & 5.56                           &  & 25.18                           & 26.77          & 22.22                           &  & \textbf{48.52}                           & \textbf{49.11}          & \textbf{47.44 }                          &  & 78.68                           & 81.48          & 73.50 \\
XML(VR + ML)~\cite{lei2020tvr}  & 18.37                           & 20.36          & 14.76                           &  & 7.92                            & 9.23           & 5.56                           &  & 27.23                           & 31.54          & 19.44                           &  & 43.81                           & 46.92          & 38.19                           &  & 81.93 & 85.38 & 75.69                           \\
DME(ours)                                             & \textbf{22.92} & \textbf{27.50} & \textbf{20.44} &  & \textbf{11.94} & \textbf{16.24} & \textbf{9.61} &  & \textbf{30.13} & \textbf{38.46} & 25.62 &  & 46.37 & 46.15 & 46.49 &  & \textbf{83.48}                           & \textbf{87.18}          & 81.48                          \\ \bottomrule[1pt]
\end{tabular}
 }
 \caption{Test set results on various methods for TQVSR. ``All'' means all queries; ``Seen'' means queries for seen scenarios; ``Unseen'' means queries for unseen scenarios.}
 \label{tab:supp_method_cmp}
\end{table*}

\section{More Details of DME Method}
\label{sec:supp_dme_method}
 Fig.~\ref{fig:model} gives an overview of DME for TQVSR. Formally, we define the inputs into the model as: a candidate video segment associated with transcripts and a user question composed of an image with or without referring regions, and a textual query. 

In our method, we represent each chunked video segment $v_i$ as a list of consecutive short clips, i.e., $v_i=\left[c_{i,1}, c_{i,2}, \dots, c_{i,l}\right]$, where $l$ is the length of video segment $v_i$ (\#clips). In AssistSR, each short clip is also associated with temporally aligned transcripts.

\noindent\head{Model architecture} 
DME is composed of two multimodal encoders based on the transformer network~\cite{vaswani2017attention}. In TQVSR, both user question and a context video segment are multimodal, i.e. both of them are composed of vision input and language input, requiring a model for learning joint contextualized representations. Inspired by the recent success of the vision and language model ~\cite{Lu2019ViLBERTPT,Su2020VL-BERT:,li2019visualbert}, we follow~\cite{li2019visualbert} to use the self-attention mechanism within the Transformer to implicitly align elements of the input text and regions in the question and model the contextual information of the input video segment.

\noindent\head{Input representations.} To represent a video segment, we consider appearance features. For each frame, we used Resnet-50~\cite{he2016deep} with weights from grid-feat~\cite{jiang2020gridfeat}, which is trained on Visual Genome~\cite{krishnavisualgenome} for object detection and attribute classification and produces effective features for image VQA tasks~\cite{antol2015vqa,gurari2018vizwiz}. We resize a frame to $448 \times 746$ and max-pool feature map after the $C5$ block to get the $2048D$ representation. We extract $2048D$ ResNet-50 features at 10 FPS and max-pool the features every 1.5 seconds to obtain a clip-level feature. To represent transcripts, we extract contextualized text features using a 6-layer pretrained DistilBert~\cite{sanh2019distilbert}. We used the implementation of DistilBert from \cite{wolf-etal-2020-hugging-face} to extract contextualized token embeddings from its second-to-last layer~\cite{lei2020tvqa+}. After extracting the token-level embedding, we then max-pool them every 1.5 seconds to get a $768D$ clip-level feature vector. We use a $768D$ zero vector if encountering no transcripts.

To represent the image and referring region in the question, we used Resnet-50~\cite{he2016deep} with weights from grid-feat~\cite{jiang2020gridfeat}. For the image, similar to video appearance, we feed it into the Resnet-50 and max-pool the feature map after the $C5$ block to obtain a $2048D$ feature. For a referring region located by a bounding box, we apply RoI pooling~\cite{girshick2014rcnn,girshick2015fast-rcnn} on the feature map after Resnet-50 $C5$ block to obtain a $2048D$ region feature.
To represent the text query in the question, we directly used the extracted token embeddings from the second-to-last layer of DistilBert~\cite{sanh2019distilbert}.

Following~\cite{li2019visualbert}, all the extracted visual features are projected into $768D$ features via a linear layer. Without ambiguity, we use the used the symbols by denoting the processed symbols as $E^{Q_t} \in \mathbb{R}^{l_q \times d}$, $E^{Q_v} \in \mathbb{R}^{N_r \times d}$, $E^{V_f} \in \mathbb{R}^{l \times d}$ and $E^{V_t} \in \mathbb{R}^{l \times d}$, where $Q_t$ represents the textual query of the question, $l_q$ is the length of this query; $Q_v$ means the visual feature of the question input, $N_r$ is the number of regions, including the whole image and regions being referred to; $V_f$ and $V_t$ means frames and transcripts of the video segment, $l$ is the number of clips in this video segment and $d$ is the hidden size ($d$ is set to be $768$ in our experiments). 

For multimodal encoders, we inject different types of input attribute into $E$ by adding two additional embedding layers: (1) \textit{token type encoding} that informs the type of information: using \texttt{[vis]} for visual features in the question and appearance feature in the video segment, while using \texttt{[txt]} for features of textual query in the question and transcript feature in video segment. (2) \textit{position encoding} that is used to inject signals of the token ordering. For textual query in the question, position encoding is following the sequence order. For appearance and transcript features in the video segment is following the order of clips sequence. As for visual features in the question, position encoding is used when alignments between words and bounding regions are provided as part of the input, and is set to the sum of the position embeddings corresponding to the aligned words as in~\cite{li2019visualbert}. 

These layers are trainable to enable models to learn the dynamics of input features and are modeled to have the same feature dimension $d$. We combine all encoding layers through element-wise summation for each modality in a multimodal encoder. Specifically for $m \in \left\{Q_v, Q_t, V_f,V_t \right\}$, the result representation is: 
$$
Z^m = E^m + E_{tok}^m + E_{pos}^m.
$$

For input representation of each encoder, we concatenate the text feature and visual feature to create a single sequence embedding:
\begin{equation}
    \begin{aligned}
        Z^Q &= \left[Z^{Q_t} ; Z^{Q_v}\right] \\ 
        Z^V &= \left[Z^{V_t} ;  Z^{V_f}\right]. \nonumber
    \end{aligned} 
\end{equation}

\noindent\head{Multimodal encoding}
Given the input features $Z^Q$, $Z^V$, as shown in Fig.~\ref{fig:model}, we use two encoders to compute their representation respectively. For each multimodal encoder, a learned \texttt{[cls]} token~\cite{Devlin2019BERTPO} is concatenated to the beginning of the input feature, which is used to produce the final  output representation of the transformer. We denote the output of the Query Encoder as $H^Q \in \mathbb{R}^d$ for the question $Q$, and the output of the Video Encoder as $H^V \in \mathbb{R}^d$ for the video segment $V$.

\noindent\head{Training and Inference} During training, one relevant segment is randomly sampled to be paired with a question. For training loss, we employ~\cite{zhai2018classification} for segment retrieval setting, where matching question-segment pairs in a batch are treated as positives, and all other pairwise combinations in the same batch are treated as negatives. We maximize the score between positive pairs and minimize the score between negative pairs.
We minimise the sum of two losses:
\begin{equation}
    \begin{aligned}
        L_1 &= -\frac{1}{B} \sum_i^{B} \log \frac{\exp\left({H_i^V}^{\top}{H_i^Q}/\sigma\right)}{\sum_{j=1}^B \exp\left({H_i^V}^{\top}{H_j^Q}/\sigma\right)} \\
        L_2 &= -\frac{1}{B} \sum_i^{B} \log \frac{\exp\left({H_i^Q}^{\top}{H_i^V}/\sigma\right)}{\sum_{j=1}^B \exp\left({H_i^Q}^{\top}{H_j^V}/\sigma\right)}, \nonumber
    \end{aligned}
\end{equation}
where $H_i^V$ and $H_j^Q$ here are the normalized embeddings of the $i$-th video segment and the $j$-th question in a batch of size $B$ and $\sigma$ is the temperature. The overall loss function is $L=L_1+L_2$ for DME.

At inference time, the DME model requires only the dot product between the multimodal question embedding and candidate video segment embeddings. This retrieval inference is of trivial cost since questions and video segments are indexable and therefore it is scalable to large scale retrieval.

Specifically, DME conducts simple dot-product at the feature-level and hence features of video segments can be pre-computed and cached.
We use the popular similarity ranking library faiss-gpu~\footnote{\url{https://github.com/facebookresearch/faiss/blob/main/tutorial/python/5-Multiple-GPUs.py}} to test the run time retrieval. We test on a server with a server with 8 RTX3090 GPUs and AMD EPYC 7413 24-Core Processor: with pre-computed video features, the retrieval time of one query is 16.65ms for a 1K video corpus and 162.47ms for a 100M video corpus. Although there is an increase in similarity ranking time, it is still fast and acceptable in practice.

\section{More Details of Experiment}
\label{sec:supp_exp}
\subsection{Details for Baseline Methods}

\textbf{\# SiameseNet}~\cite{chopra2005learning} for image-video matching. As a simple baseline, we use the SiameseNet to match the query image and the frames of the answer segments. This SiameseNet baseline does not include any textual inputs (question and transcripts).

\noindent\textbf{\# TVQA.}~\cite{lei2018tvqa} proposed a multi-stream end-to-end trainable neural network for Multi-Modal VideoQA. In this model, the question-answer pairs are used to fuse with visual features and text features in the video separately. We modify the input module to fit in our multimodal question, and fuse the question with different modalities from the paired video segment separately.

\noindent\textbf{\# XML}~\cite{lei2020tvr} is a late fusion approach for Video Corpus Moment Retrieval (VCMR). In XML, separated self-cross-encoders are used to encode visual and textual features of a video. The query is feeded into a self-attention module followed by a FC-layer to generate modularized query representations. Late fusion is then applied for Video Retrieval and moment retrieval. We add a module image query in AQVSR, which is the same to the original query branch in XML. We use element-wise adding for image query and text query, to generate modularized query. XML(VR) views each video segment in AQVSR as a video, and just compiles the Video Retrieval part in XML. XML(VR+ML) uses the rough start-end time in AssistSR annotation for VCMR setting. At inference time, we pick the highest score of the moments in video segments as the score for that video segment; for each query, we rank the scores of all candidate video segment.

\noindent\textbf{\# ClipBERT}~\cite{lei2021less} is a generic framework for video-language tasks. We modify ClipBERT a bit to fit in our multi-modal query: we linearly project the  ResNet-50~\cite{he2016deep}(pretrained weights from~\cite{jiang2020gridfeat}) $C5$ max-pooled feature of the image query to text query feature. We also concatenate corresponding transcript features for sampled frames to the transformer encoder used in ClipBert~\cite{lei2021less} for visual-textual fusion.

\noindent\subsection{Implementation Details}
 We use the same set of offline extracted features for DME, TVQA~\cite{lei2018tvqa} and XML~\cite{lei2020tvr}. We use ResNet-50~\cite{he2016deep} pretrained by~\cite{jiang2020gridfeat} as feature extractor for image query and use a 6-layer DistilBert~\cite{sanh2019distilbert} for text feature extraction.  For all transformer-based encoders, we set the number of hidden layers to be 4 and set hidden size to be 768. In the baseline experiment, we do not use any pretrained model for the transformer encoder for fair comparison.
 For ClipBERT~\cite{lei2021less}, we follow the text-video retrival protocol in the original implementation. We employed $4 \times 1$ (randomly sample 4 video clips and randomly sample 1 frame within each clip) during training. We sampled 16 video clips and the middle frame of each clip during testing. The duration of each video clip is 1.5 seconds, which is the same in other baselines.
We keep other settings the same as the original implementation for 
TVQA\footnote{\url{https://github.com/jayleicn/TVQA}}, XML\footnote{\url{https://github.com/jayleicn/TVRetrieval}} and ClipBERT\footnote{\url{https://github.com/jayleicn/ClipBERT}}. 
For DME, We use Adam~\cite{Kingma2015AdamAM} optimizer and a learning rate of $3 \times 10^{-5}$. The hyper-parameter $\sigma$ for the training loss is set to be $0.05$. 
 
 \subsection{More Experimental Results on AssistSR}
 In Tab.~\ref{tab:supp_method_cmp}, we show the performance (mAP/Recall@1/Recall@5/Recall@10/Recall@50) of different methods on AssistSR. We show the performance on all questions, seen scenarios (related to 73.7\% quesitons) and unseen scenarios (related to 26.3\% questions).
 We can observe that (1) DME clearly
outperforms other methods for video-related tasks, indicating that models designed for other tasks can not generalize well to the TQVSR task. (2) For DME, ClipBERT~\cite{lei2021less} and XML~\cite{lei2020tvr}, performance on seen scenarios is better than that on unseen scenarios, indicating that for these methods, they are able to capture something in common within the same scenario but fail to generalize well to unseen scenarios. This is because for items within the same scenario could share similar
knowledge, such as structure and functionality, while for items for totally different scenarios,
models may fail to bridge the gap of the difference in appearance, language style, etc.

\subsection{Ablation for input fusion type.} 
\begin{table}[h]
\resizebox{.98\linewidth}{!}{
\begin{tabular}{l|cccc}
\hline
Method               & mAP   & R@1    & R@5    & R@10   \\ \hline
Adding Fusion        & 17.86 & 7.21  & 26.88 & 40.49 \\
Concatenating Fusion & 22.92 & 11.94 & 30.13 & 46.37 \\ \hline
\end{tabular}
}

\caption{Results for different input fusion types for DME.}
\label{tab:exp_fusion_type}
\end{table}

We study the effect of different fusion types for the input of DME model. Here we conduct experiments on two fusion types for the multimodal encoder. (1) Adding fusion. For each multimodal encoder in DME, fuse visual and textual features of a video with positional alignment via element-wise adding as in~\cite{Su2020VL-BERT:}; token type embedding is not adopted in this case. (2) Concatenating fusion. For each multimodal encoder in DME, we concatenate the textual feature and visual feature into a single sequence embedding. This is used for our DME baseline. Experiment results in Tab.~\ref{tab:exp_fusion_type} show that applying concatenating fusion yields better performance over adding fusion. This suggests that to give full play to the advantages of Transformer in multimodal reasoning, it is necessary to decouple multi-modal information into separate sequences.

\end{document}